%% file: paper.tex
\documentclass[]{article}

\usepackage{soul, amssymb, amsmath, xfrac, float}
\usepackage[usenames, dvipsnames]{color}
\usepackage[hidelinks]{hyperref}
\usepackage[dvipsnames]{xcolor}
\usepackage[linesnumbered, ruled, vlined]{algorithm2e}

\usepackage{xcolor}

\usepackage{subcaption, pifont}
\newcommand{\xmark}{\ding{55}}

\usepackage[top=2cm, bottom=2cm, left=2cm, right=2cm]{geometry}

\begin{document}


\title{Trajectory Clustering Performance Evaluation: If we know the answer, it's not clustering}

%


\author{Mohsen Rezaie\\
Polytechnique Montr\'eal\\
Montr\'eal, Qu\'ebec, Canada
\and
Nicolas Saunier\\
Polytechnique Montr\'eal\\
Montr\'eal, Qu\'ebec, Canada}


\maketitle

\input{9-0-abstract.tex}

\paragraph{Keywords} trajectory clustering, unsupervised evaluation, similarity measure, traffic pattern recognition


\input{9-1-intro.tex}
\input{9-2-related_works.tex}
\input{9-3-method.tex}
\input{9-4-dataset.tex}

\input{9-5-results.tex}
\input{9-6-conc.tex}

\bibliographystyle{plain} 
\bibliography{paper}

\end{document}

%% file: 9-0-abstract.tex
\begin{abstract}

Advancements in Intelligent Traffic Systems (ITS) have made huge amounts of traffic data available through automatic data collection. A big part of this data is stored as trajectories of moving vehicles and road users. Automatic analysis of this data with minimal human supervision would both lower the costs and eliminate subjectivity of the analysis. 
Trajectory clustering is an unsupervised task. 
It requires a measure to calculate similarity of the trajectories, a clustering algorithm to extract the patterns and a measure to evaluate its performance.

In this paper, we perform a comprehensive comparison of similarity measures, clustering algorithms and evaluation measures using trajectory data from seven intersections. We also propose a method to automatically generate trajectory reference clusters based on their origin and destination points to be used for label-based evaluation measures. Therefore, the entire procedure remains unsupervised both in clustering and evaluation levels. Finally, we use a combination of evaluation measures to find the top performing similarity measures and clustering algorithms for each intersection. The results show that there is no single combination of distance and clustering algorithm that is always among the top ten clustering setups. 


\end{abstract}

%% file: 9-1-intro.tex
\section{Introduction}\label{ch:paper_introduction}

Data collection and analysis is the basis for transport planning and traffic management. Better serving users while minimizing the negative impacts of transportation systems requires better data. From automatic traffic counting, surveys, incident detection, road network state recognition to law enforcement tasks, transport management tasks all benefit from increasing amounts of data with good temporal and spatial coverage collected thanks to the advancements in hardware and software technology~\cite{bommes2016video}. Exploiting the full potential of the massive data collected every day by Intelligent Transportation Systems (ITS) requires automatic data management and analysis with minimum intervention from human operators.

Whether the data collection be through fixed or moving sensors, e.g.\ on vehicles or pedestrians, such as pole or drone mounted regular or infrared cameras, radars, LIDARs and Global Navigation Satellite System (GNSS) devices~\cite{barmpounakis2020new, antoniou2011synthesis, haghani2010freeway, bernas2018survey}, various kinds of sensors provide user trajectories, i.e.\ the series of their positions over time. These trajectories are then processed in various ways for many transport applications, as noted above. 


In some data analysis tasks such as motion prediction, the expected answer of the model for each trajectory is known, either retrospectively (for example in the case of time series prediction) or through expert annotation. Such a task is therefore supervised, i.e.\ a label is available for each trajectory. If the label is a discrete variable, it is a classification task, while if it is continuous, it is regression task. 
However, this is typically not the case for many other tasks such as automatically learning OD matrices, identifying trip patterns in the population, anomaly detection for example to spot incidents or law violations
~\cite{moreira2016time, bandaragoda2019trajectory, kwon2013scene, zhen2017maritime, rezaie2017semi, katrakazas2015real}. Such tasks consist in putting together similar trajectories into subsets of the whole dataset using a similarity measure or distance, also called clustering. These tasks are naturally unsupervised, as a good partition of the dataset or clustering depends on the application or purpose. Even when possible, annotation is time consuming, costly, error-prone and can be subjective or ambiguous. 

We refer to the combination of clustering algorithm, distance or similarity metric and their parameters as the clustering setup. Finally, performance measures are used to evaluate how well the clustering method is working.

Even though clustering algorithms are unsupervised by definition, a large body of literature use labeled data to evaluate the performance of these models. 
Since the performance of a given clustering setup may depend on each dataset it is applied to, this approach means that some data must be labeled every time to measure the performance on new data, which is impractical and negates to some extent the need for clustering in the first place. This motivates our exploration of unsupervised performance measures for trajectory clustering. 


In this paper we propose a framework for a comprehensive search of suitable trajectory-clustering setups with common unsupervised evaluation measures. We also propose a method to generate reference clusters for trajectories based on their origin and destination points to be used in supervised evaluation measures. We then use the results from the performance measures (both unsupervised and supervised) and combine them to generate a metric to compare several popular trajectory distances and clustering algorithms. 
The remainder of this paper is organized as follows. Section~\ref{ch:paper_related_work} outlines the prior literature related to trajectory clustering and discusses how this research builds on the prior work. Section~\ref{ch:paper_methodology} quickly introduces the similarity measures, clustering algorithms and evaluation methods and then explains the steps of the proposed comprehensive comparison and evaluation method. 
Section~\ref{ch:paper_results} discusses the results and insights from testing the clustering algorithm on real-world data. Section~\ref{ch:paper_conclusion} describes the conclusions of this research and the proposed future work.

%% file: 9-2-related_works.tex
\section{Related Work}\label{ch:paper_related_work}


Trajectory-based road user detection, classification and tracking has been deployed for different traffic problems in various environments such as road user counting, speed measurement and movement detection (e.g.\ traffic violation detection), road user interaction analysis and automatic incident detection \cite{buch2011review, kan2019traffic, rezaie2017semi, wang2017automatic}.





A crucial part of time-series data clustering problems is choosing a suitable measure of similarity or distance. There are two categories of similarity measures used in time-series analysis: lock-step and elastic measures~\cite{ding2008querying, ferreira2016time}. ``Lock-step measures compare fixed (one-to-one) pairs of elements''~\cite{ferreira2016time} from input time series of the same length: well-known measures are the Bhattacharyya distance, the Euclidean Distance (ED) and more generally the distances derived from the $L_p$ norms. On the other hand, elastic measures such as Dynamic Time Warping (DTW), distances based on Longest Common Sub-Sequence (LCSS), Edit distance with Real Penalty (ERP), Edit Distance on Real sequence (EDR) and Symmetric Segment-Path Distance (SSPD) can handle time-series of different lengths. Among these similarity measures, LCSS and DTW are very popular among researchers for their ability to compare trajectories with different lengths~\cite{aghabozorgi2015time, morris2009learning, saunier2007probabilistic, besse2016review}. However, this flexibility in input length usually comes at the cost of higher computational burden. For instance, the computational burden for the ED and Bhattacharyya distances is $O(n)$ while computing several elastic measures including DTW and LCSS are of $O(mn)$ where $m$ and $n$ is the lengths of the input time-series~\cite{bian2018survey}.

A body of literature focuses on trajectory clustering, using different Machine Learning (ML) approaches. Choong et al.\ used trajectory clustering for the traffic surveillance at intersections~\cite{choong2016vehicle}. They used Longest Common Sub-Sequence (LCSS) for measuring similarity and then used Gaussian kernel function to convert LCSS similarity into distance. K-means and Fuzzy C-Means (FCM) are used for clustering and the results were compared against ground-truth through the Rand Index (RI).

Fu et al.\ used average pair-wise distance of first \textit{N} points of each pair of trajectories to represent the dissimilarity of them, where \textit{N} is the length of the shorter trajectory~\cite{fu2005similarity}. Then they used two layer clustering based on FCM, Spectral and hierarchical clustering algorithms and compared them against each other. For evaluation of performances they used Tightness and Separation Criterion (TSC) which is based on within and between cluster dispersion of objects.


Zhao et.al \cite{zhao2019trajectory} used Douglas-Peucker (DP) based compression to re-sample marine trajectory data. DTW is then used to calculate distances between the trajectories. Finally Density-Based Spatial Clustering of Applications with Noise (DBSCAN) algorithm is used for clustering. A custom performance metric based on ground-truth labels was used for evaluation of the method.

Belisle et al.~worked on automatically counting vehicle movements at intersections~\cite{belisle2017optimized}. Their method consists of vehicle detection and tracking, then trajectory clustering  with a focus on parameter optimization. They used LCSS as similarity measure and a customized clustering algorithm similar to K-means. The algorithm picks trajectory prototypes to represent each cluster. For optimization and evaluation purposes, traffic turning counts were used as ground truth to calculate the Weighted Average Percentage Error (WAPE).



Even though different methods of similarity measurement and clustering algorithms are applied in trajectory analysis and novel methods have been purposed, unsupervised performance measurement is mostly overlooked. Given the heuristic nature of clustering approaches and the concern about their sensitivity to differing data characteristics, errors in data and also in judgement~\cite{milligan1987methodology}, reiterating evaluation measurements for each site could be essential.

In this study, we focus on providing a framework to use several internal and external performance indices without ground-truth labels to facilitate easier clustering evaluations. More specifically, we propose a method to generate reference labels for supervised performance metrics and use a combination of supervised and unsupervised performance measures.

%% file: 9-3-method.tex
\section{Methodology}\label{ch:paper_methodology}
\subsection{Clustering Setup}
For each site of interest, let $D$ be the set of road user trajectories within the boundaries of the site. Let each trajectory $M_i \in D$ be a time-ordered sequence of $m_i$ time-stamped positions of road user $i$ as $(x^i_j,y^i_j,t^i_j)$ where $x^i_j$ and $y^i_j$ represent the coordinates of the road user at timestamp $t^i_j$, ${1\leq j \leq m_i}$. 

Clustering algorithms split the dataset into groups with similar objects based on a similarity or distance measure. According to the literature, we use the terms "distance" and "dissimilarity" in their broad meaning to mention a method that measures unlikeness of a pair of trajectories. Value of a distance measure is essentially non-negative and the more alike a pair of trajectories are, the lower is the distance value. On the other hand, a similarity measure represents a method of comparing trajectories which assigns (still non-negative) higher values when two trajectories are more alike~\cite{chen2009similarity}.

\begin{table}[h!]
    \centering
    \begin{tabular}{l|l|l}
         (Dis)similarity Measures & Parameters & Experimental Values \\
         \hline
         DTW & - & - \\
         LCSS & $r_b$ &  1, 2, 3, 5, 7, 10 m\\
         EDR & $r_b$ &  1, 2, 3, 5, 7, 10 m\\
         PF & $w$ & 0.01, 0.05, 0.1, 0.2, 0.3, 0.5\\
         Hausdorff & - & -\\
         SSPD & - & -
    \end{tabular}
    \caption{Similarity measures}
    \label{tab:similariy_measures}
\end{table}

Table~\ref{tab:similariy_measures} shows an overview of the (dis)similarity measures used in this study and their parameters.
For all the measures in Table~\ref{tab:similariy_measures}, the Euclidean distance was used for calculating point-to-point distances. LCSS and EDR both have a parameter $r_b$ which is the threshold on point distances to decide if two points are close enough and may be part of their similar subsequence. 

Since LCSS provides a similarity measure,  we convert it to the distance $d_{LCSS}$ as suggested by Vlachos et al.~\cite{vlachos2002discovering}:

\begin{equation}
    d_{LCSS}(M_i,M_j) = \frac{1-LCSS(M_i,M_j)}{min(m_i,m_j)}
    \label{eq:lcss-conversion}
\end{equation}

where $LCSS(M_i,M_j)$ is the LCSS similarity between trajectories $M_i$ and $M_j$, i.e.\ the lengths of the longest common subsequence between them.

Given a distance measure $d_\alpha$ with parameters $\alpha$, all distances $d_{i,j}$ between all pairs of trajectories $<M_i, M_j>$ in $D$ are calculated and stored in matrix $T_{D, d_\alpha}$. 
The distance matrix $T_{D, d_\alpha}$ is then used by a clustering algorithm $A^1$ with parameters $\beta$ to split $D$ into $k$ groups. We define the clustering setup as $\omega=(d, \alpha, A^1, \beta, k)$. Even though there are algorithms which do not use the whole matrix but rather some of its elements (e.g.\ DBSCAN and spectral clustering), pre-computing the matrix saves time since it is used repeatedly in our experiments for different clustering algorithms $A^1$ and number of clusters $k$.

In this study, we used six popular clustering algorithms used for trajectory clustering, namely,
k-Medoids~\cite{park2009simple}, agglomerative hierarchical clustering~\cite{jain1999data}, spectral clustering~\cite{shi2000normalized}, Density-based spatial clustering of applications with noise (DBSCAN) and Ordering Points To Identify The Clustering Structure (OPTICS).

It is worth mentioning that DBSCAN and OPTICS do not take into account a number of clusters as input. In these two cases, the number of clusters $k$ is a result of the algorithm and cannot be controlled directly. 
The table~\ref{tab:clustering_algorithms} summarizes the clustering algorithms used in this study with their corresponding parameters (in addition to the number of clusters when possible).

\begin{table}[h!]
    \centering
    \begin{tabular}{l|l}
         Clustering Algorithm & Parameters \\
         \hline
         k-medoids & - \\
         agglomerative clustering & $linkage\in \{complete, average, single\}$ \\
         spectral & - \\
         DBSCAN & $n_z^{min}$, $d_z$ \\
         OPTICS & $n_z^{min}$
    \end{tabular}
    \caption{Clustering algorithms}
    \label{tab:clustering_algorithms}
\end{table}

\begin{figure}[!ht]
\centering
\centering
  \includegraphics[width=0.7\linewidth]{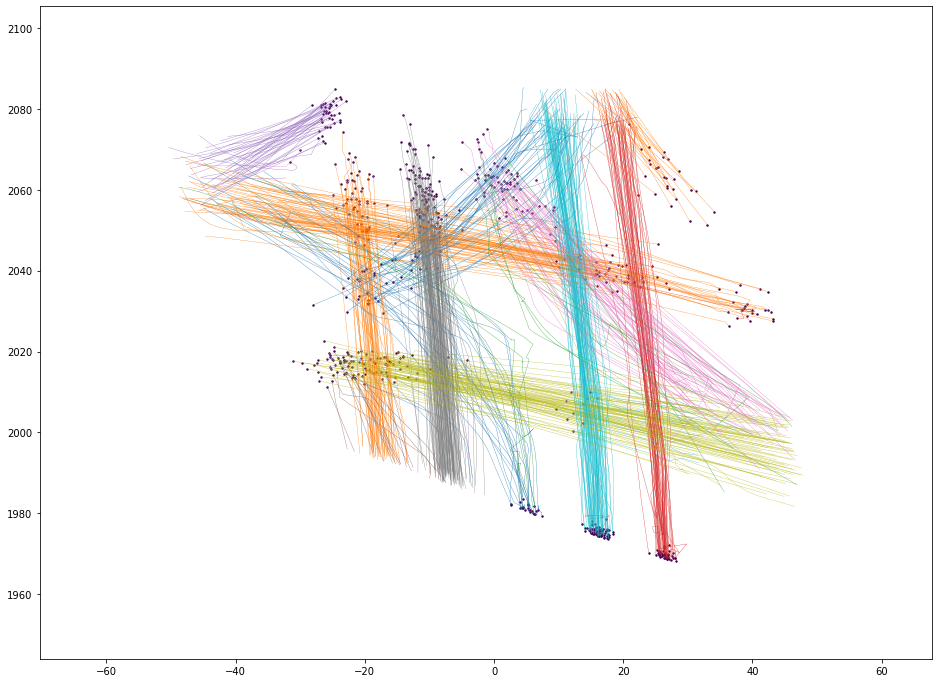}
  \caption{Example of trajectories $D$ in an intersection after clustering. Clusters are represented by different colors (in some cases the colors assigned to different clusters are similar due to the limit on number of colors while maintaining the contrast high enough to keep them visually distinctive).}
  \label{fig:traj_NGSIM5_clusters_together_2020-10-12}
\end{figure}

Figure~\ref{fig:traj_NGSIM5_clusters_together_2020-10-12} shows an example of trajectories recorded from an intersection in Atlanta, Georgia, after trajectory clustering.

For the reasons highlighted in the introduction of cost, time and subjectivity of manually labelling trajectory clusters, the focus of this work is on the unsupervised evaluation of the performance of methods for trajectory clustering. Besides, there is an inherent contradiction to the use of labels for clustering: if labels can be unambiguously assigned, it means the task can be supervised and is not a clustering task.

Thus, there are no reference to check the clustering results presented in Figure~\ref{fig:traj_NGSIM5_clusters_together_2020-10-12}. While we may visually confirm that the generated clusters are meaningful and can be considered as representatives of different manoeuvres in the intersection, this is not always the case. 
The clusters presented in Figure~\ref{fig:traj_NGSIM5_clusters_together_2020-10-12} result from a clustering setup $\omega$ picked based on the results of this work. Indeed, it is not trivial to make the choice of $\omega$ and the best clustering setup may be different from one location or period to another because of different characteristics of the road configuration and the road user trajectories. The next sections introduce the performance metrics used in this work and how supervised performance metrics can be used without ground truth. 


\subsection{Unsupervised Performance Measure}
There are several unsupervised performance metrics for clustering evaluation. These metrics only rely on distance values to evaluate the clustering setup $\omega$. The silhouette (S)~\cite{rousseeuw1987silhouettes} is one of these metrics which is widely used in the literature for trajectory clustering evaluation~\cite{aghabozorgi2015time}. It is defined for each object $i$ as~\cite{rousseeuw1987silhouettes}:

\begin{equation}
    s_i = \frac{b_i-a_i}{max(a_i,b_i)}
\end{equation}

Where $a_i$ is the mean distance between the object $i$ and all other objects within the same cluster, and $b_i$ is the smallest of the mean distances between the object $i$ and the objects in another cluster (different from the one object $i$ belongs to). The total clustering performance based on the Silhouette $S$ is then calculated as the average silhouette coefficient over all objects.

Other unsupervised performance measures like he Calinski-Harabasz Index (CHI) \cite{maulik2002performance} and the Davies-Bouldin Index (DBI)~\cite{davies1979cluster} cannot be used as they rely on the computation of cluster centers as averages of all objects in each cluster, which cannot be done simply for variable length vectors such as trajectories. 

However, unsupervised performance measures  such as Silhouette are mostly suitable for comparison between clustering setups with the same algorithms since they make assumptions about cluster structure~\cite{aghabozorgi2015time}. Also, unsupervised performance measures cannot reflect results expected for specific applications. If an expected, even partial, reference clustering is known, generating a result that matches this clustering may not yield the best unsupervised performance measure.

\begin{figure*}[!ht]
\centering
\begin{subfigure}{0.45\linewidth}
\centering
  \includegraphics[width=\linewidth]{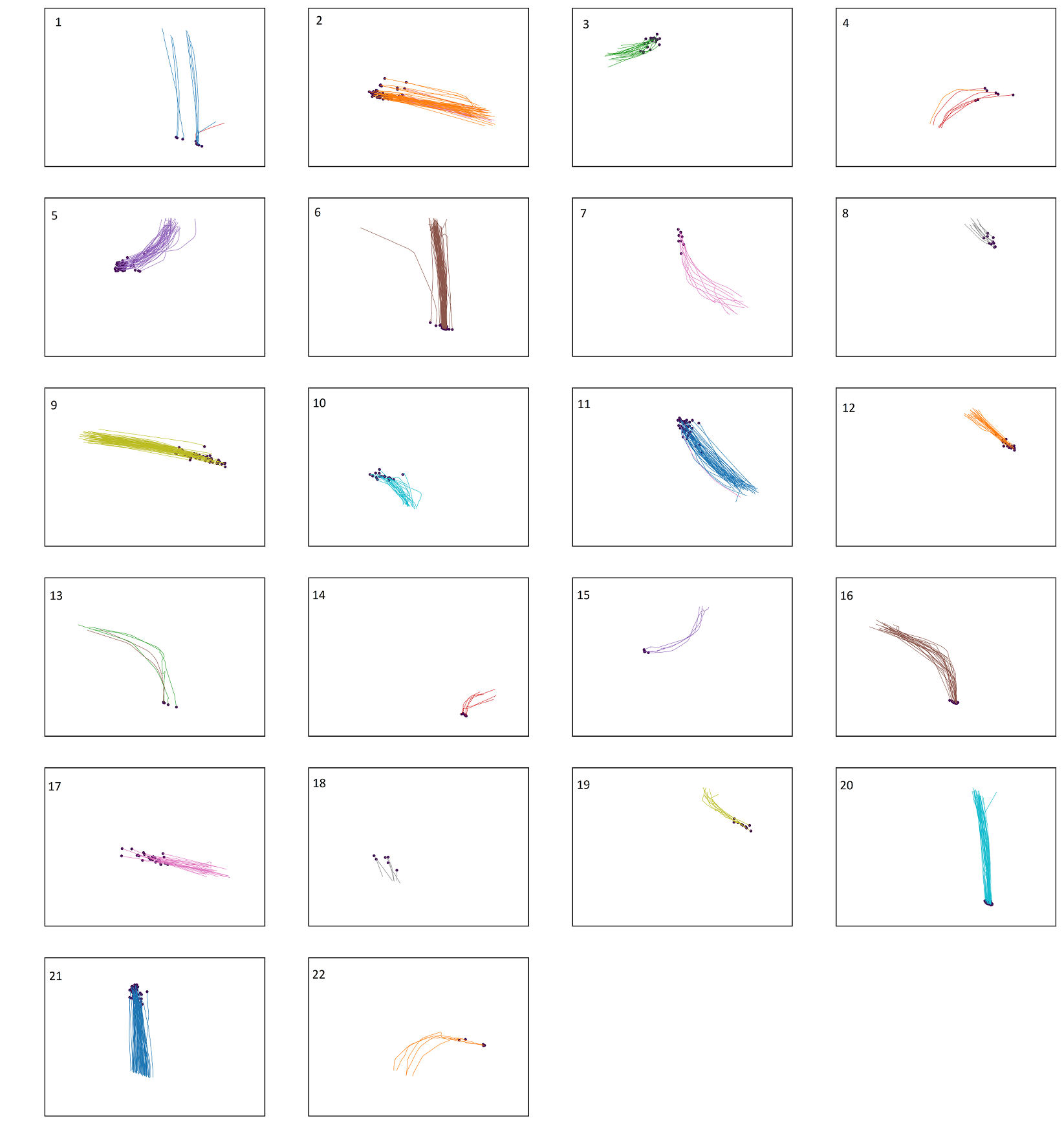}
  \caption{Clustered trajectories with colors showing the closest cluster for each trajectory, computed as the mean distance to all elements in that cluster.}
  \label{fig:traj_NGSIM5_clusters_2020-10-09}
\end{subfigure}
\hspace{0.5cm}
\begin{subfigure}{0.45\linewidth}
\centering
  \includegraphics[width=\linewidth]{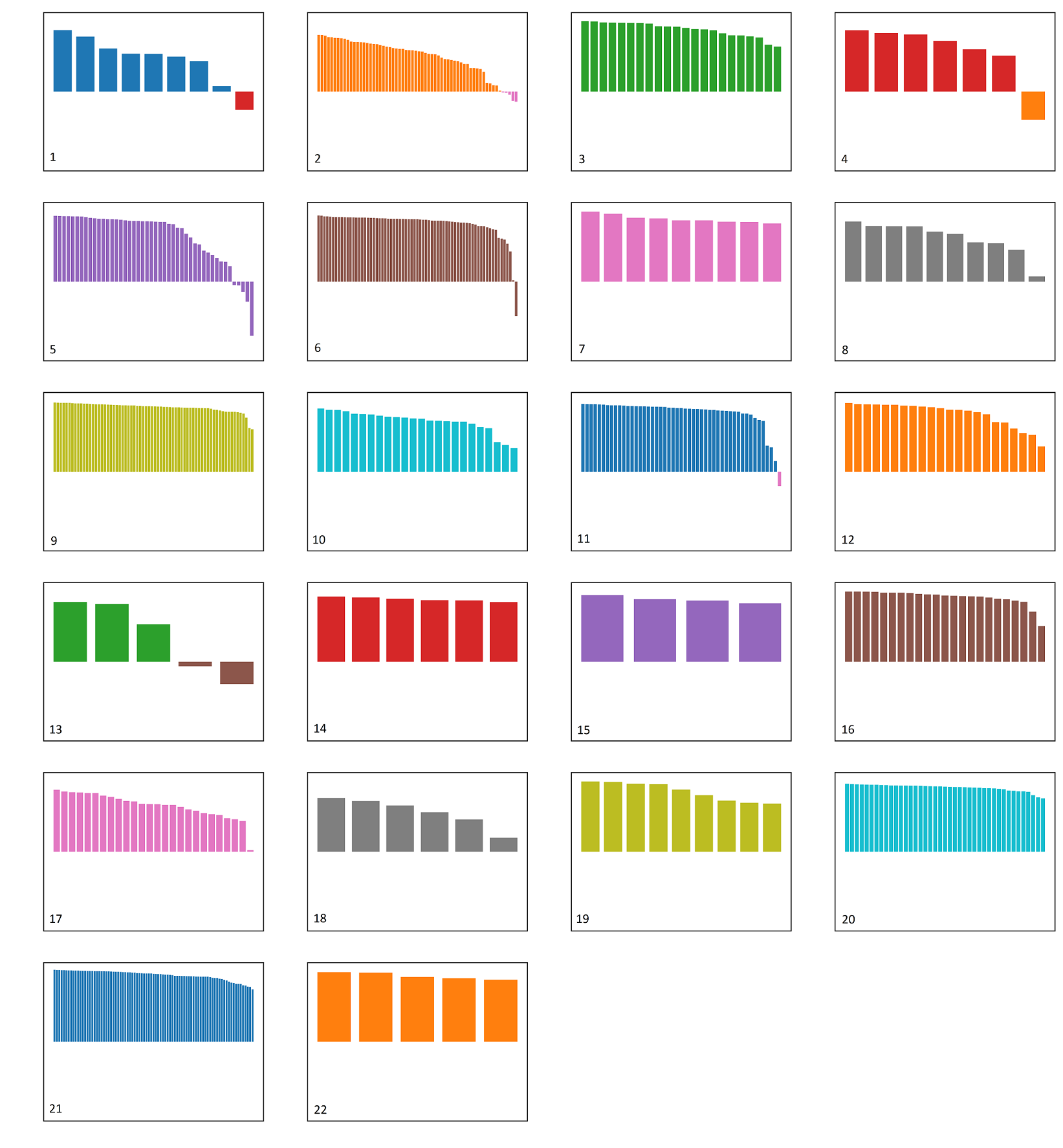}
  \caption{Silhouette values for trajectories in the clusters shown on the left}
  \label{fig:traj_NGSIM5_sil_2020-10-09}
\end{subfigure}
\caption{Example of error in clustered trajectories and their silhouette values}
\label{fig:traj_NGSIM5_sil_error}
\end{figure*}

We have observed that if two trajectories are mistakenly considered similar by $A^1$ based on distance $d$, an unsupervised performance measure based on $d$ may make the same mistake and confirm $A^1$'s decision. Figure~\ref{fig:traj_NGSIM5_sil_error} shows an example case in which the silhouette method mistakenly confirms the clustering results. As one can see, the cluster 1 should be split in two groups, each being merged with either cluster 6 or 20. Thus, while we expect several negative silhouette values for the trajectories in the first cluster, they are mostly positive and confirm the clustering mistakes.

\subsection{Weak Trajectory Clustering Supervision using Origin-Destination}
To overcome this problem, we suggest that beside using common unsupervised evaluation measures, a method independent of the distance $d$ and trajectory clustering algorithm $A^1$ be used to validate the result of trajectory clustering. The method that we propose here is generating ``reference'' trajectory clusters based on clustering simply the first and last points of the trajectories (origins and destinations) and using these reference clusters with supervised evaluation measures. 

To do so, we extract the origins and destinations of the trajectories in $D$ and put them in two sets, respectively $D_O$ and $D_D$. The points in each set are clustered using the Euclidean distance and the clustering algorithm $A^2$, which may not be the same as $A^1$, with the parameters $\beta_O$ and $\beta_D$ and number of clusters $k_O$ and $k_D$, respectively. The steps are provided in detail in Algorithm~\ref{algo:od}.

\begin{figure}[!ht]
  \centering
  \includegraphics[width=\linewidth]{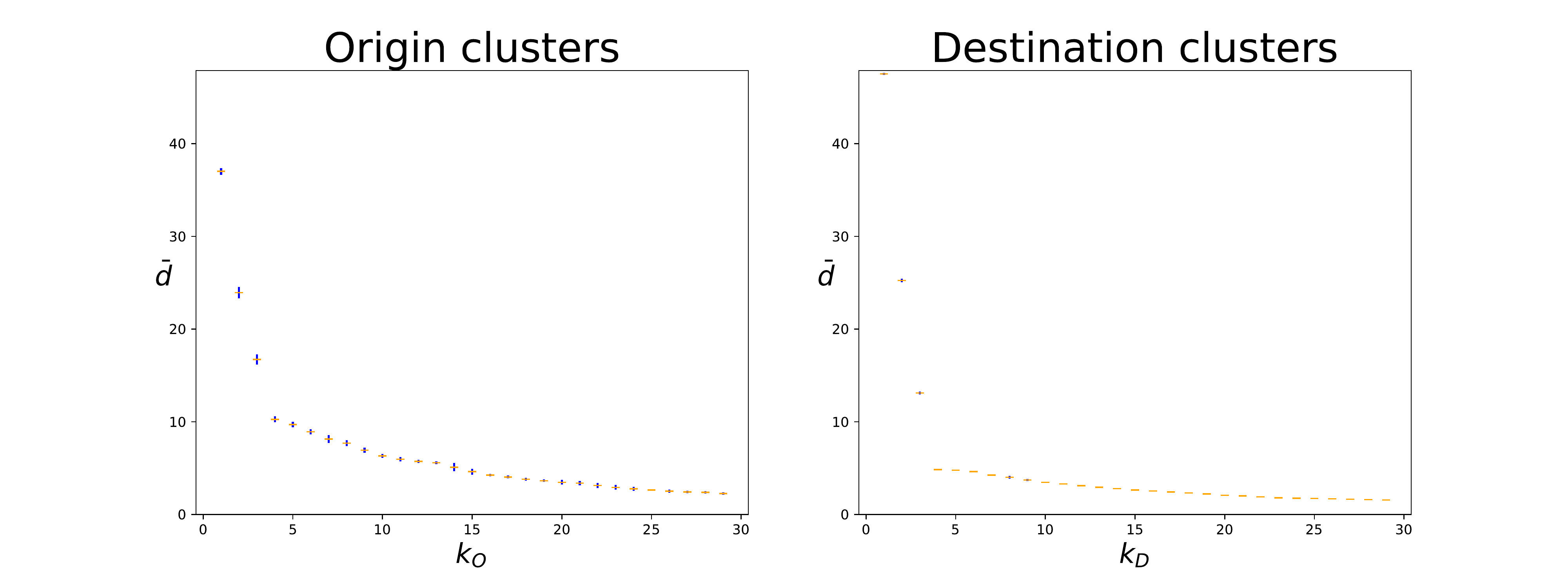}
  \caption{Average distances for the datasets of origins (left) and destinations (right) with respect to the number of clusters. The mean plus and minus two standard deviations are computed over several replications of the clustering algorithm for each number of clusters.
  }
  \label{fig:NGSIM-OD-number_of_clusters}
\end{figure}

\begin{figure}[!ht]
  \centering
      \includegraphics[width=0.24\textwidth]{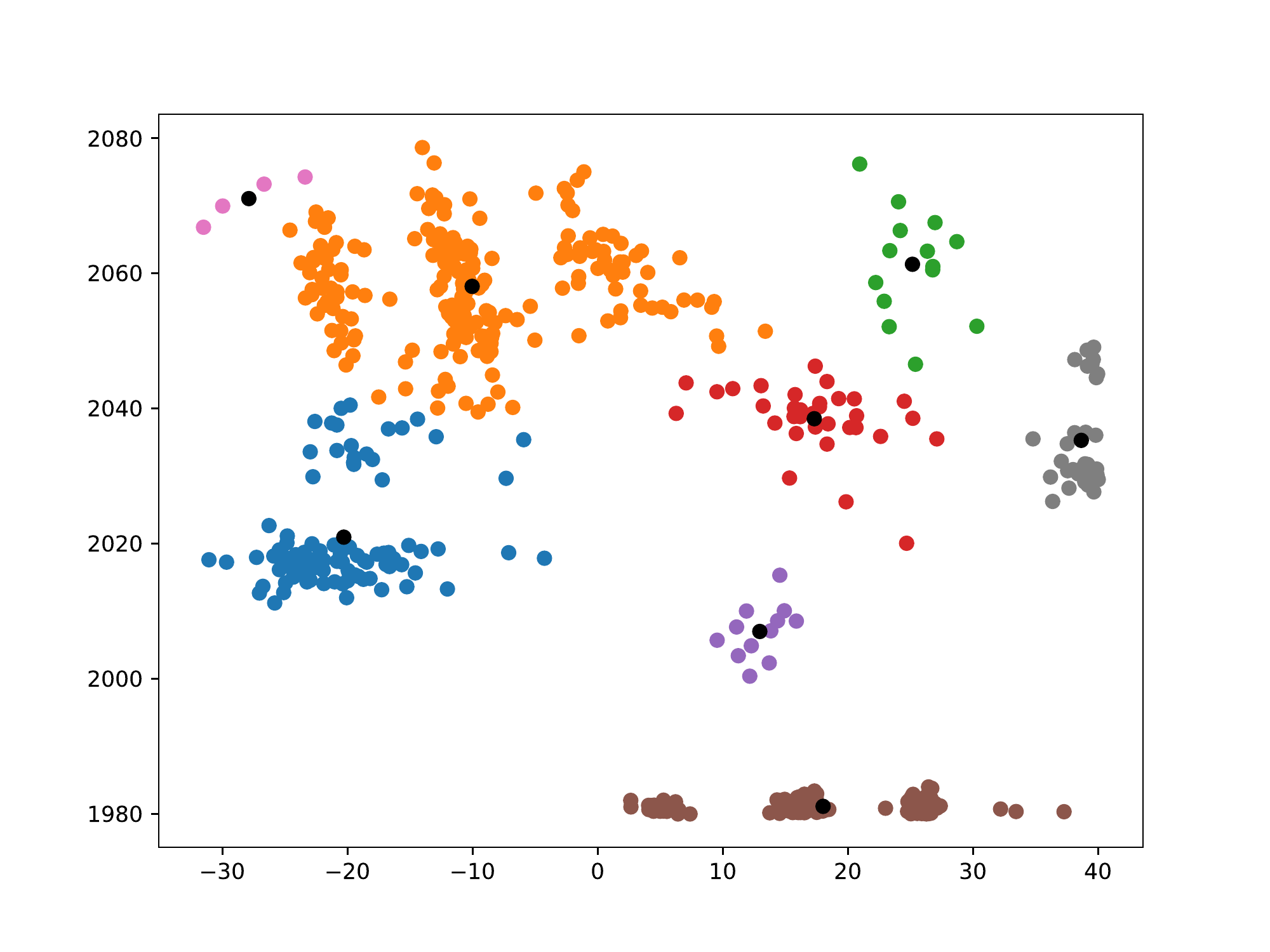}\includegraphics[width=0.24\textwidth]{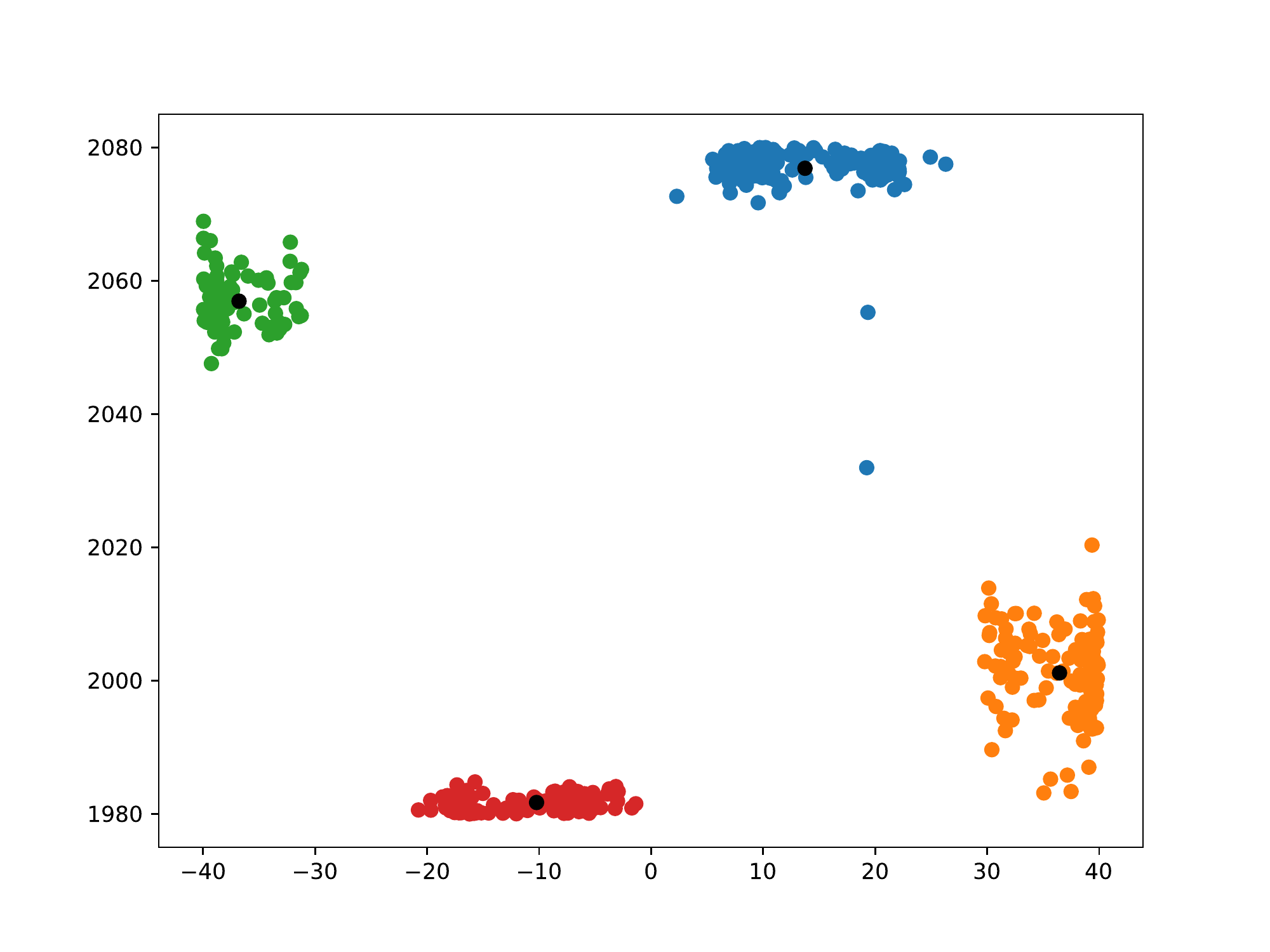}
  \caption{Trajectory origin (left) and destination (right) clusters with $k_O=8$, $k_D=4$ and agglomerative hierarchical clustering algorithm with average linkage.}
  \label{fig:od_NGSIM5_clusters_2020-10-12}
\end{figure}

The elbow method is used to make good choices for $k_O$ and $k_D$, since it is a popular technique in point clustering~\cite{kodinariya2013review}. Figure~\ref{fig:NGSIM-OD-number_of_clusters} illustrates the average distances $\bar{d}$ with respect to $k_O$ and $k_D$: in this case we picked $k_O=8$ and $k_D=4$. The graphs are produced based on data from an intersection in NGSIM dataset using the agglomerative hierarchical clustering algorithm with average linkage as the merging criterion. Figure~\ref{fig:od_NGSIM5_clusters_2020-10-12} depicts an example of the origin and destination clusters for the trajectories in the same intersection.
If the proportion of trajectories starting from origin cluster $\phi_O$ and destination cluster $\phi_D$ is more than a threshold $\epsilon=0.01$, i.e.\ over 1~\% of all trajectories, the origin-destination pair is deemed significant and we assign label $\phi_{OD}$ to the corresponding trajectories.

Using these reference clusters, we calculate six additional performance measures, namely, the completeness, homogeneity and Validity (V) measures~\cite{rosenberg2007v}, the Adjusted Rand Index (ARI)~\cite{santos2009use}, Adjusted Mutual Information (AMI)~\cite{vinh2010information} and Fowlkes-Mallows Index (FMI)~\cite{desgraupes2013clustering}. These measures represent the agreement or similarity between the reference clusters and the clusters $K$ generated by the given setup $\omega$, ignoring permutations of the clusters. In particular, for ARI and FMI, the measures rely on the joint classifications of object pairs, whether they are both in the same or different clusters in the reference and $K$. 

\begin{itemize}
\item \textbf{Completeness, homogeneity and V}~\cite{rosenberg2007v}: for a cluster assignment $K$, we define entropy of $K$:

\begin{equation}
    H(K) = -\sum^{k}_{i=1} \frac{|C_i|}{n}.\log{\frac{|C_i|}{n}}
\end{equation}

The entropy $H(Ref)$ of the reference clusters $Ref$ is defined accordingly. Let $k_{Ref}$ be the number of reference clusters. The conditional entropy is also defined, for example of $Ref$ given $K$ $H(Ref|K)$:

\begin{equation}
    H(Ref|K) = -\sum^{k_{Ref}}_{r=1} \sum^{k}_{i=1} \frac{n_{r,i}}{n}.\log{\frac{n_{r,i}}{n_i}}
\end{equation}

Where $n_{r,i}$ is the number of objects belonging to reference cluster $r$ and cluster $i$. The conditional entropy $H(K|Ref)$ of $K$ given $Ref$ is defined in a symmetric manner. We can then finally define the performance measures completeness ($c$) and homogeneity ($h$), and $V$ as their harmonic mean:

\begin{align}
    c &= 1 - \frac{H(K|Ref)}{H(K)}\\
    h &= 1 - \frac{H(Ref|K)}{H(Ref)}\\
    V &= \frac{(1+\beta) \times h \times c}{\beta \times h + c}
\end{align}

We use $\beta=1$ to give equal weights to the completeness and homogeneity measures.

\item \textbf{Adjusted Mutual Information (AMI)}: we first define the probabilities $P(r)=\frac{|Ref_r|}{n}$ and $P'(i)=\frac{|C_i|}{n}$ of a randomly picked object to belong respectively to reference cluster $r$ and cluster $i$, and $P(r,i)=\frac{n_{r,i}}{n}$. We then define the Mutual Information (MI)~\cite{vinh2010information}:
\begin{equation}
    MI(Ref,K) = \sum^{k_{Ref}}_{r=1} \sum^{k}_{i=1} P(r,i)\log{\frac{P(r,i)}{P(r)P'(i)}}
\end{equation}

Since the MI favors clustering with a higher number of clusters $k$, we use the Adjusted Mutual Information (AMI) instead:

\begin{equation}
    AMI = \frac{MI-E[MI]}{mean(H(Ref),H(K))-E[MI]}
\end{equation}

Where $E[MI]$ is the expected value of the MI index~\cite{vinh2010information}.

\item the \textbf{Adjusted Rand Index (ARI)} is a simple, yet popular index used for external clustering performance. It is defined as~\cite{santos2009use}:

\begin{equation}
    RI = \frac{a+b}{C^n_2}
\end{equation}

Where $a$ is the number of object pairs that are in same cluster in both $Ref$ and $K$, $b$ is the number of object pairs that are in different clusters in both $Ref$ and $K$, and $C^n_2=\frac{n(n-1)}{2}$ is the total number of object pairs in the dataset. For a random label assignment, RI is not necessarily zero. Instead, we use the Adjusted Rand Index (ARI) which fixes this issue using the expected value of RI $E[RI]$:

\begin{equation}
    ARI = \frac{RI - E[RI]}{max(RI)-E[RI]}
\end{equation}

\item the \textbf{Fowlkes-Mallows Index (FMI)} is also based on the classification of the object pairs, considering two objects being in the same clusters in both assignments as the positive class and two objects being in different ones as the negative class. For such a binary classification problem, the numbers of True Positives (TP), False Positives (FP) and False Negatives (FN) can be counted and the Fowlkes-Mallows Index (FMI) is calculated as the geometric mean of precision and recall:~\cite{desgraupes2013clustering}:

\begin{equation}
    FMI = \frac{TP}{\sqrt{(TP+FP)(TP+FN)}}
\end{equation}

\end{itemize}

Table~\ref{tab:evaluation_metrics} provides an overview of the performance measures used in this work.

\begin{table}[h!]
    \centering
    \begin{tabular}{l|c|c|c}
         \shortstack{Evaluation\\Measures} & \shortstack{Need for\\Reference Clusters} & Higher-the-better & Range \\
         \hline
         S & \xmark & \checkmark & $[-1,1]$ \\
         completeness & \checkmark & \checkmark & $[0,1]$ \\
         homogeneity & \checkmark & \checkmark & $[0,1]$ \\
         V & \checkmark & \checkmark & $[0,1]$ \\
         AMI & \checkmark & \checkmark & $[-1,1]$ \\
         ARI & \checkmark & \checkmark & $[-1,1]$ \\
         FMI & \checkmark & \checkmark & $[0,1]$ 
    \end{tabular}
    \caption{Evaluation metrics}
    \label{tab:evaluation_metrics}
\end{table}

The trajectory dataset was permuted $n=10$ times in order to make the results more robust for each clustering setup $\omega$.
For each clustering setup, the mean and standard deviation are computed for each performance measure, to compute intervals of the performance measures as the mean plus or minus two standard deviations which are plotted in the result section. 


Finally, we use a combination of these evaluation measures to identify top performing clustering setups. For each performance measure $v$ and each $\omega$ we calculate the average performance $\bar{v_\omega}$, and, 
assuming t-student distributions (unknown standard deviation), we calculate the lower bound of the confidence interval for 95~\% confidence level $v^{lower}_\omega$:

\begin{equation}
    v^{lower}_\omega = \bar{v}_\omega - t_{95\%, 9} * \frac{\hat{\sigma}_{v,\omega}}{\sqrt{n}}
\end{equation}

Then for a given evaluation measure $v$, we use $v^{lower}_\omega$ 
to order $\omega$ and assign rank $r_{v,\omega}$ to each setup $\omega$. Finally these rankings are averaged over different evaluation measures $v$ to calculate the combined performance of each clustering setup $\bar{r_\omega}$. A formal description of the computation of the clustering performance is presented in Algorithm~\ref{algo:trajectory_clustering}.

\begin{algorithm}[!ht]
    \SetKwInOut{Input}{input}\SetKwInOut{Output}{output}
    \Input{Sets of trajectories $D$ with derived sets of origins $D_O$ and destinations $D_D$, minimum and maximum number of clusters $k_{min}$ and $k_{max}$, cluster size threshold $\epsilon$, clustering algorithm $A^2$}
    \Output{Set $\Phi_{OD}$ of trajectory clusters $\phi_{OD}$}
    \For{$k_O \in [k_{min},k_{max})$}{
        Cluster trajectories based on their origins using $A^2$ into $k_O$ clusters\; 
        Let $\Phi_O$ be the set of all clusters $\phi_{O_i}$\ with average center $c_{O_i}$\;
        Calculate the average within-cluster sum of squares or average distance $\bar{d}_{k_O}=\frac{ \sum\limits^{k_O}_{i=1}{ \sum_{p \in \phi_{O_i}} ||p-c_{O_i}|| }}{|D|}$ using the $L_2$ norm $||\ ||$\;
        }
    Draw the graph of $\bar{d}_{k_O}$ values against $k_O$\;
    Pick the $\tilde{k}_O$ where an unusual drop in the $\bar{d}$ values occurs\;
    Repeat all the steps above for $D_D$ and pick $\tilde{k}_D$\;
    Split $D$ into $\tilde{k}_{OD} = \tilde{k}_O*\tilde{k}_D$ clusters depending on the cluster of the origin and destination of each trajectory\;
    \Return the set $\Phi_{OD}$ of all trajectory clusters $\phi_{OD}$ containing at least $\epsilon$ trajectories\;
\caption{Clustering trajectories based on their origins and destinations to produce reference clusters}
\label{algo:od}
\end{algorithm}

\begin{algorithm}[!ht]
    \SetKwInOut{Input}{input}\SetKwInOut{Output}{output}
    \Input{Sets of unsupervised and supervised performance evaluation measures $V^u$ and $V^s$, number of iterations $n$, subset of trajectories $\tilde{D} = \cup_{\phi_{OD} \in \Phi_{OD}} \phi_{OD}$}
    \Output{Average performance rank for the clustering setups $\bar{r_\omega}$}
    \For{each candidate $d_\alpha$}{
        Calculate distance matrix $T_{D,d_\alpha}$ with elements $d_{i,j}=d_\alpha(M_i, M_j)$\;
        \For{$l \in [1,n]$}{
            Permute randomly $\tilde{D}$\; 
            \For{each candidate clustering algorithms $A^1$, parameters $\beta$ and number of clusters $k$}{
                Cluster $\tilde{D}$ with the setup $\omega=(d,\alpha,A^1,\beta,k)$ \; 
                Let $\Phi_l$ be the set of resulting clusters\;
                For all $v \in V^u$ calculate performance measure $v_{\omega,l} = v(\Phi_l)$\;
                For all $v \in V^s$ calculate performance measure  $v_{\omega,l}=v(\Phi_l,\Phi_{OD})$\;
                }
            }
        }
    \For{$v \in V^u\cup V^s$}{
        \For{each $\omega=(s,\alpha,A^1,\beta,k)\in \Omega$}{
            Calculate average performance measure $\bar{v}_\omega = \frac{ \sum^{n}_{l=1}v_{\omega,l}(\Phi_i)
            }{n_{iter}}$\;
            Calculate lower bound $v^{lower}_\omega = \bar{v}_\omega - t_{95\%, 9} * \hat{\sigma}_{v,\omega}$\;
            }
        Order all clustering setups $\omega$ based on $v^{lower}_\omega$ and assign rank $r_{v,\omega}$\;
    }
        \For{$\omega \in \Omega$}{
            Calculate average rank $\bar{r}_\omega = \frac{\sum_{v \in V^u\cup V^s}{r_{v,\omega}}} {|V^u\cup V^s|}$
        }
    
\caption{Computing the clustering performance measures with permutation of the trajectory dataset}
\label{algo:trajectory_clustering}
\end{algorithm}

%% file: 9-4-dataset.tex


%% file: 9-5-results.tex
\section{Experimental Results}\label{ch:paper_results}
All the codes developed in this work to produce the results are available online under an open source license in a \href{https://github.com/Mr28/Poly-MSc-thesis-trajectory-clustering}{GitHub repository}.

\subsection{Datasets}\label{sec:paper_datasets}
We have have tested our framework on data from seven intersections from two datasets:

\subsubsection{NGSIM}
The first dataset used in this work is extracted from a dataset made publicly available by the Next Generation SIMulation (NGSIM) program~\cite{halkias2006next}. It is a popular dataset containing trajectory data collected at four sites in the U.S. and used among traffic researchers for benchmarking their works. It consists of positions of vehicles moving on or crossing the five lane Peachtree street in Atlanta, Georgia
. The data is recorded at 10~Hz during both the morning and afternoon rush-hours from multiple synced cameras mounted on tall buildings to cover the roads over several hundreds meters. The (sub-)trajectories within the boundaries of four intersections were extracted from this dataset. Figure~\ref{fig:NGSIM1-site_map} shows an aerial photo of intersection NGSIM 1.


\subsubsection{inD}
\begin{figure}[htbp]
    \begin{center}
	\begin{subfigure}{0.66\linewidth}
	  \includegraphics[width=\linewidth]{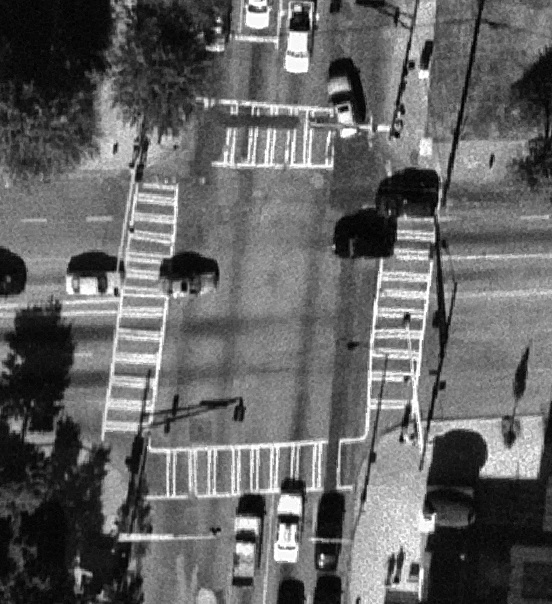}    
        \caption{Site map of intersection NGSIM 1 (one out of four)}
        \label{fig:NGSIM1-site_map}
	\end{subfigure}
	\begin{subfigure}{0.66\linewidth}
	  \includegraphics[width=\linewidth]{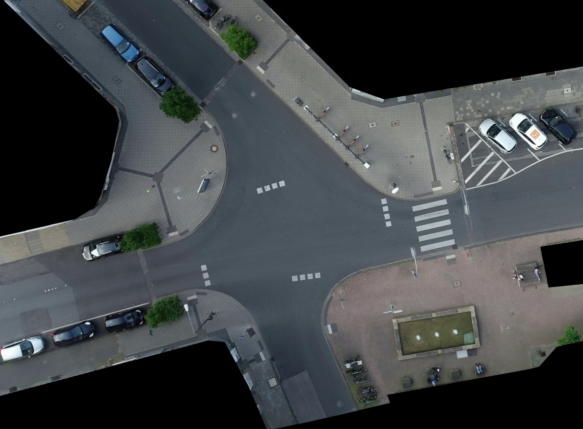}    
        \caption{Site map of intersection inD 2 (one out of three)}
        \label{fig:inD2-site_map}
	\end{subfigure}
    \end{center}
    \caption{An example of intersection for each of the two datasets}
\end{figure}

The second set of trajectories is from the intersection Drone (inD) dataset, collected from three intersections in Aachen, Germany from 2017 to 2019~\cite{bock2019ind}. The trajectories are extracted from video recordings of around 20 minutes collected via a drone-mounted camera at 25~Hz frame rate. Figure~\ref{fig:inD2-site_map} shows an aerial photo of intersection inD~2.

The table~\ref{tab:data} gives a summary of sites used for the experimental results.

\begin{table}[h!]
    \centering
    \begin{tabular}{l|l|c|c}
         Site & Camera & Intersection Types & Traffic Lights \\
         \hline
         inD 1 & drone & 4-way & \xmark \\
         inD 2 & drone & 4-way & \xmark \\
         inD 3 & drone & T-intersection & \xmark \\
         NGSIM 1 & building & 4-way & \checkmark \\
         NGSIM 2 & building & 4-way & \checkmark \\
         NGSIM 3 & building & 4-way & \checkmark \\
         NGSIM 4 & building & 4-way & \checkmark
    \end{tabular}
    \caption{Site summary with their characteristics (multiple synced cameras were mounted on buildings for the NGSIM datasets).}
    \label{tab:data}
\end{table}

\subsection{Performance Plots}
\begin{figure*}[!ht]
  \centering
  \includegraphics[width=\textwidth]{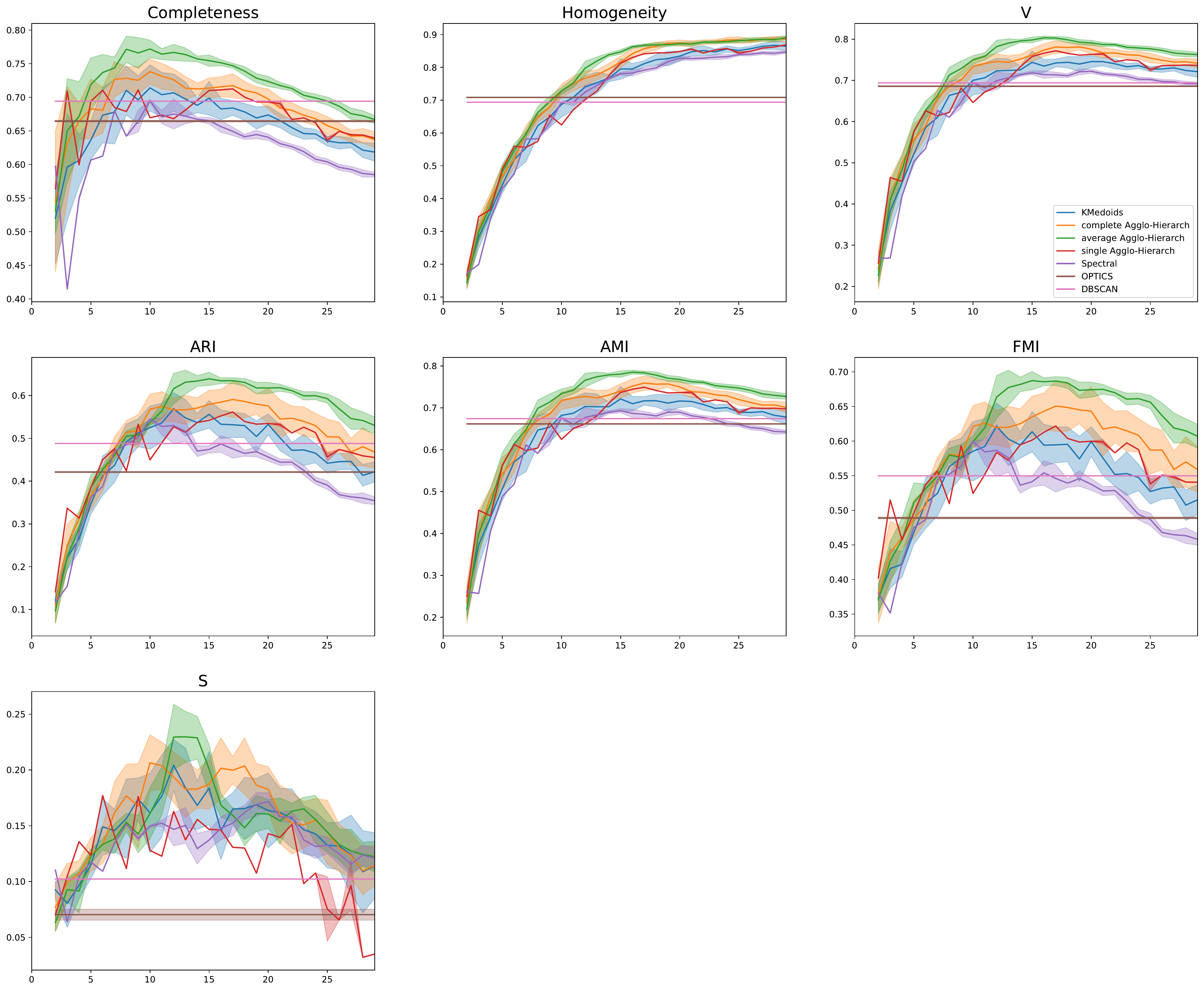}
  \caption{Performance results for the \textit{NGSIM 1} intersection based on EDR similarity measure with parameter $w = 7m$}
  \label{fig:NGSIM1_performances_Edr_param3}
\end{figure*}

Figure~\ref{fig:NGSIM1_performances_Edr_param3} shows the performance results for \textit{NGSIM 1} intersection based on the EDR similarity measure with parameter $w = 7$. Each of the plots in the figure shows one of the seven performance measures as a function of the number of clusters $k$ used in the clustering. 

Given seven intersections studied in this work, three distances $d$ with no parameters and distances having a parameter with six tested parameter values $\alpha$, there are $7*21=147$ figures similar to Figure~\ref{fig:NGSIM1_performances_Edr_param3}. Given that the general patterns observed here in Figure~\ref{fig:NGSIM1_performances_Edr_param3} is not necessarily shared in the other 146 figures, analyzing such enormous amount of data and finding the best clustering setup $\omega=(d, \alpha, A^1, \beta, n_{A^1}, k)$ is not feasible through visual comparison of those 147 figures.

\subsection{Performance Measure Combination}
\begin{figure}[!ht]
  \centering
  \includegraphics[width=\linewidth]{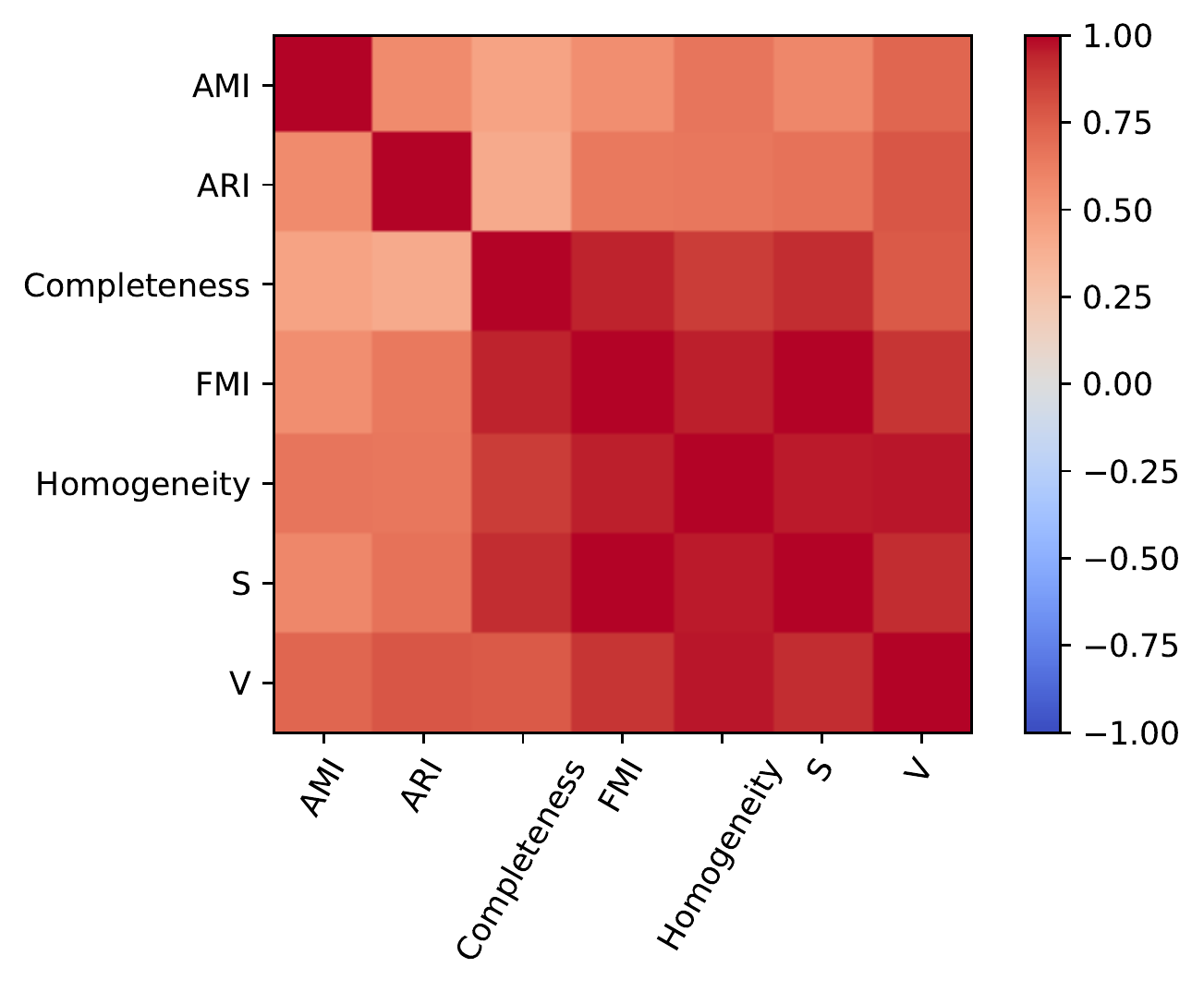}
  \caption{Correlation of the performance measures for the NGSIM~1 intersection}
  \label{fig:NGSI1_evalCorrelations}
\end{figure}

To summarize the information and ease the decision process on picking the best setup, we want to combine the performance results. However, some evaluation measures may be correlated with each other and including all of them regardless of the correlations would over-emphasize the performance features corresponding to these correlated measures. Figure~\ref{fig:NGSI1_evalCorrelations} shows the correlations between the seven performance measures for the NGSIM ~1 intersection.

Based on correlation plots from the seven intersections, evaluation measures V and homogeneity on the one hand, FMI and S on the other hand are highly correlated together with absolute correlation values between them above 0.75 for all pairs of measures in all intersections.
Therefore, we keep use S, Completeness, Homogeneity, ARI and AMI for the combined metric. 

\begin{figure}[htbp]
\begin{center}
\begin{subfigure}{0.4\linewidth}
  \centering
  \includegraphics[width=.9\linewidth]{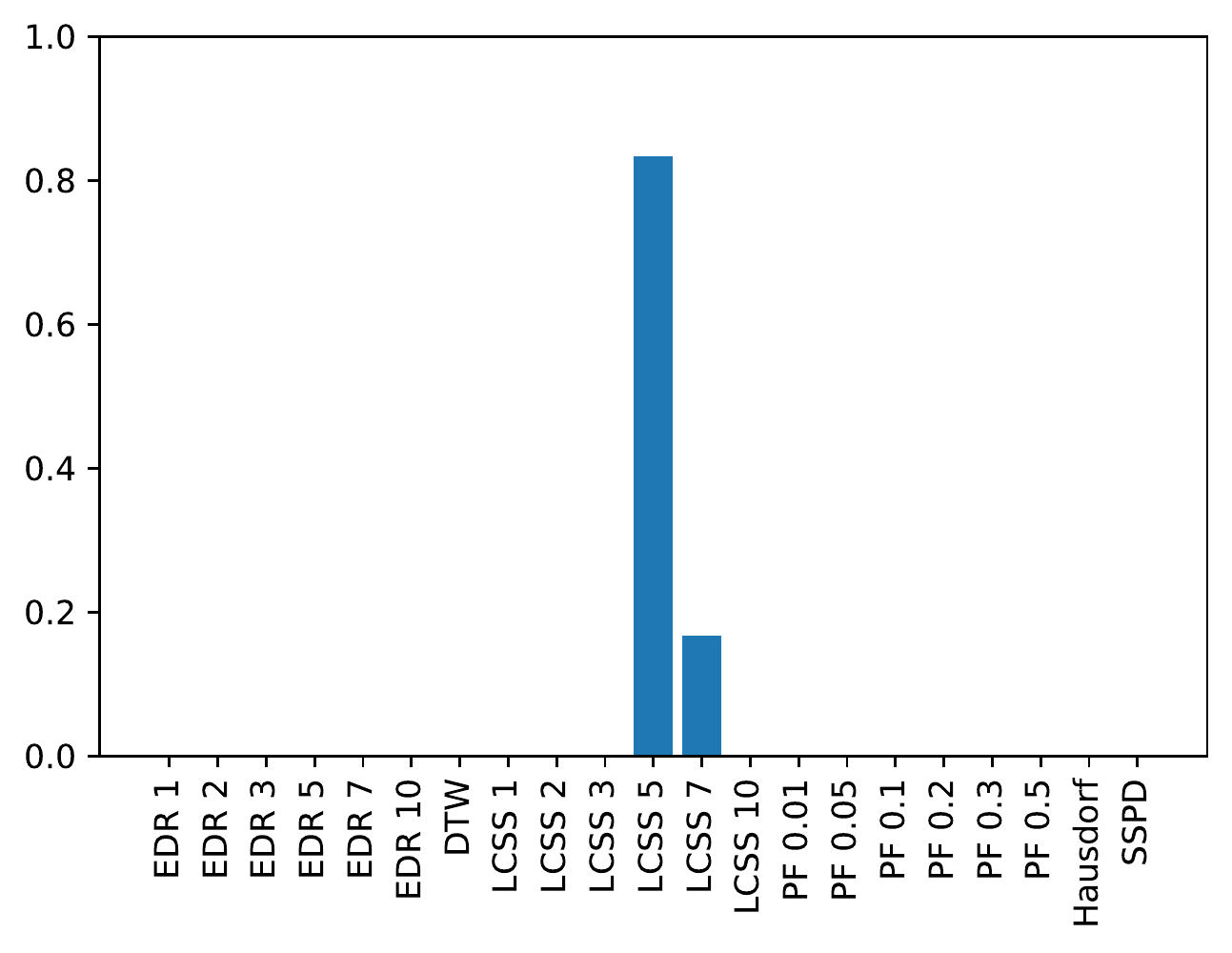}
  \label{fig:inD2_top_freq_simMeasures}
\end{subfigure}
\begin{subfigure}{0.4\linewidth}
  \centering
  \includegraphics[width=.9\linewidth]{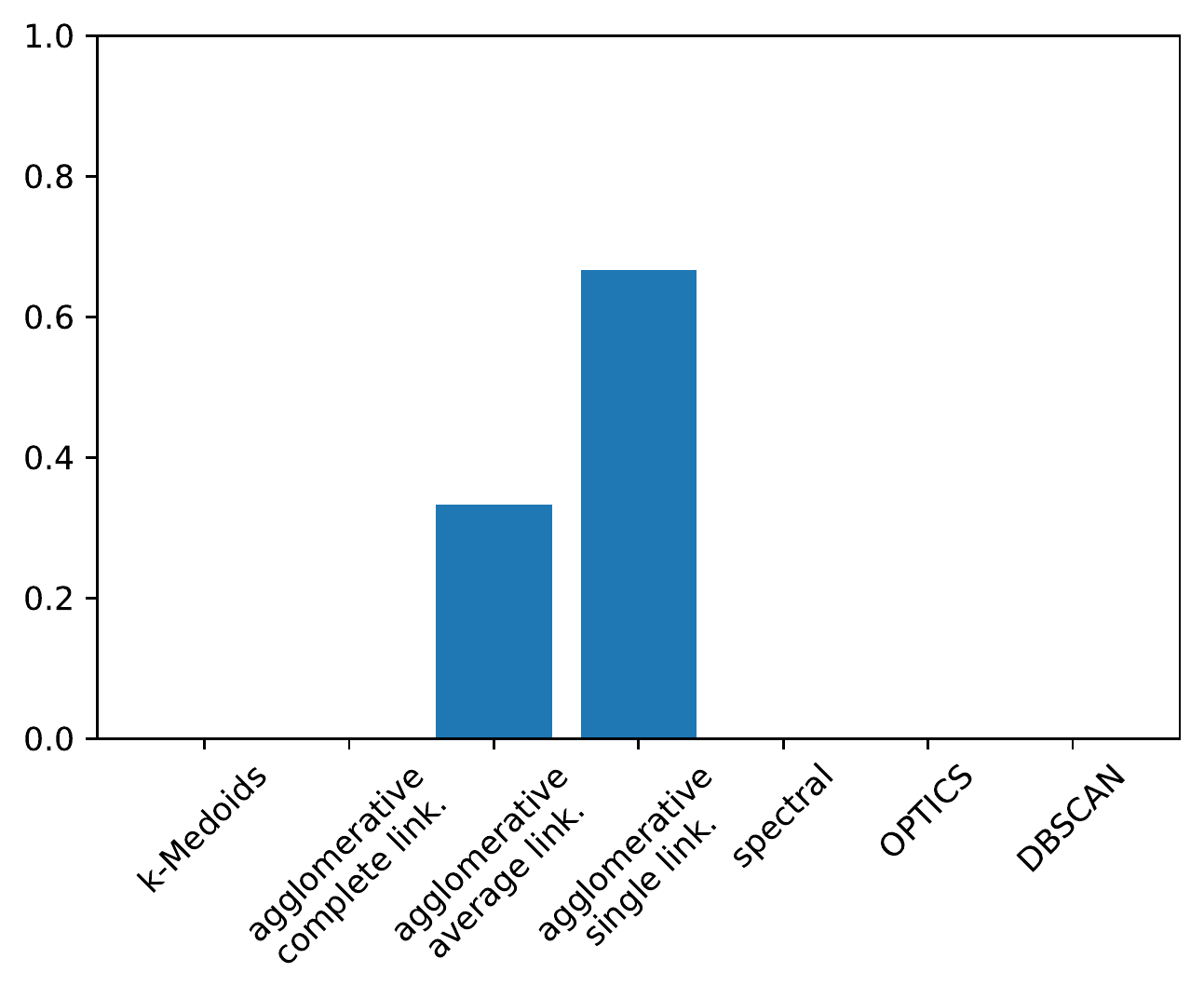}
  \label{fig:inD2_top_freq_algs}
\end{subfigure}
\caption*{inD intersection 1}

\begin{subfigure}{0.4\linewidth}
  \centering
  \includegraphics[width=.9\linewidth]{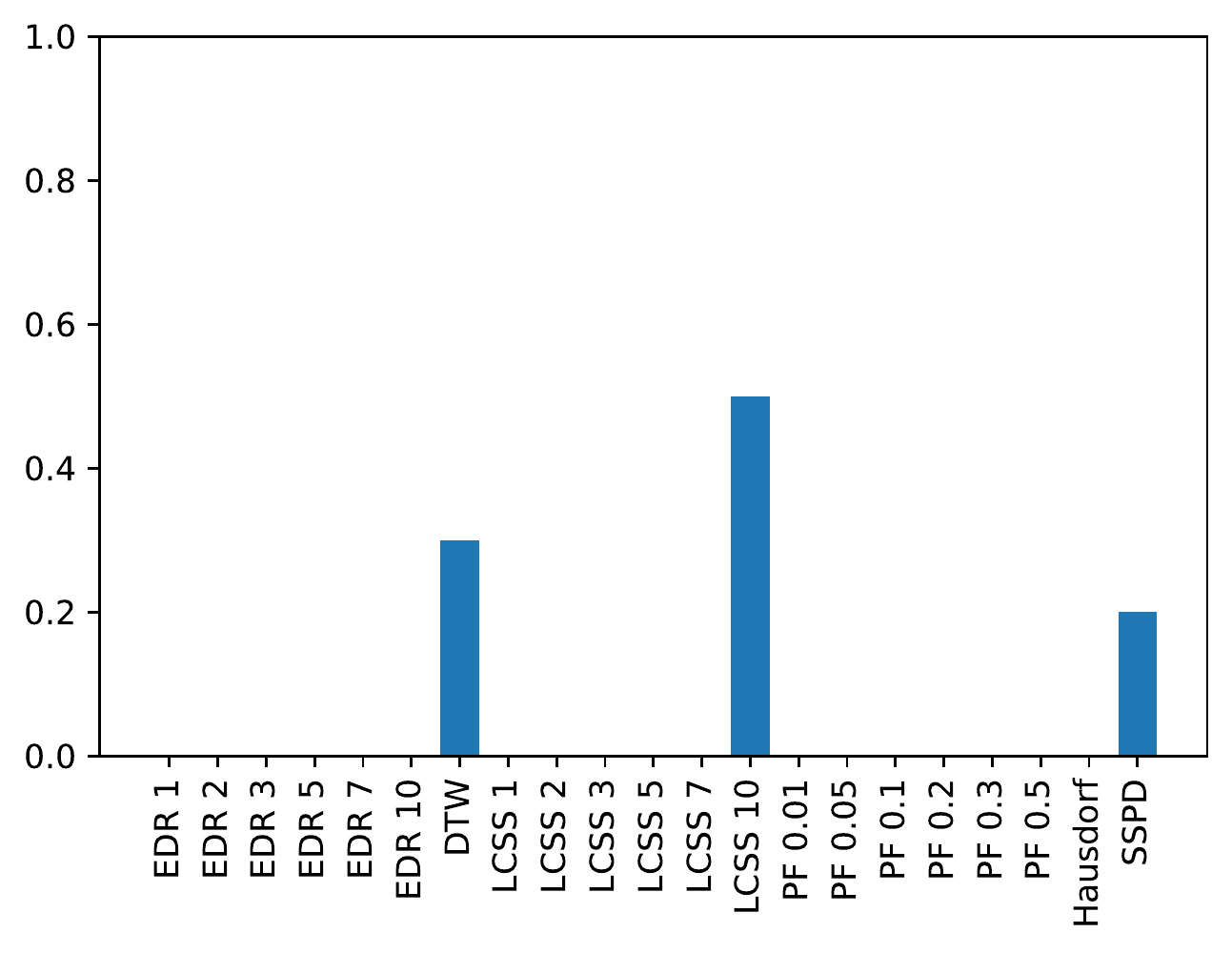}
  \label{fig:inD3_top_freq_simMeasures}
\end{subfigure}
\begin{subfigure}{0.4\linewidth}
  \centering
  \includegraphics[width=.9\linewidth]{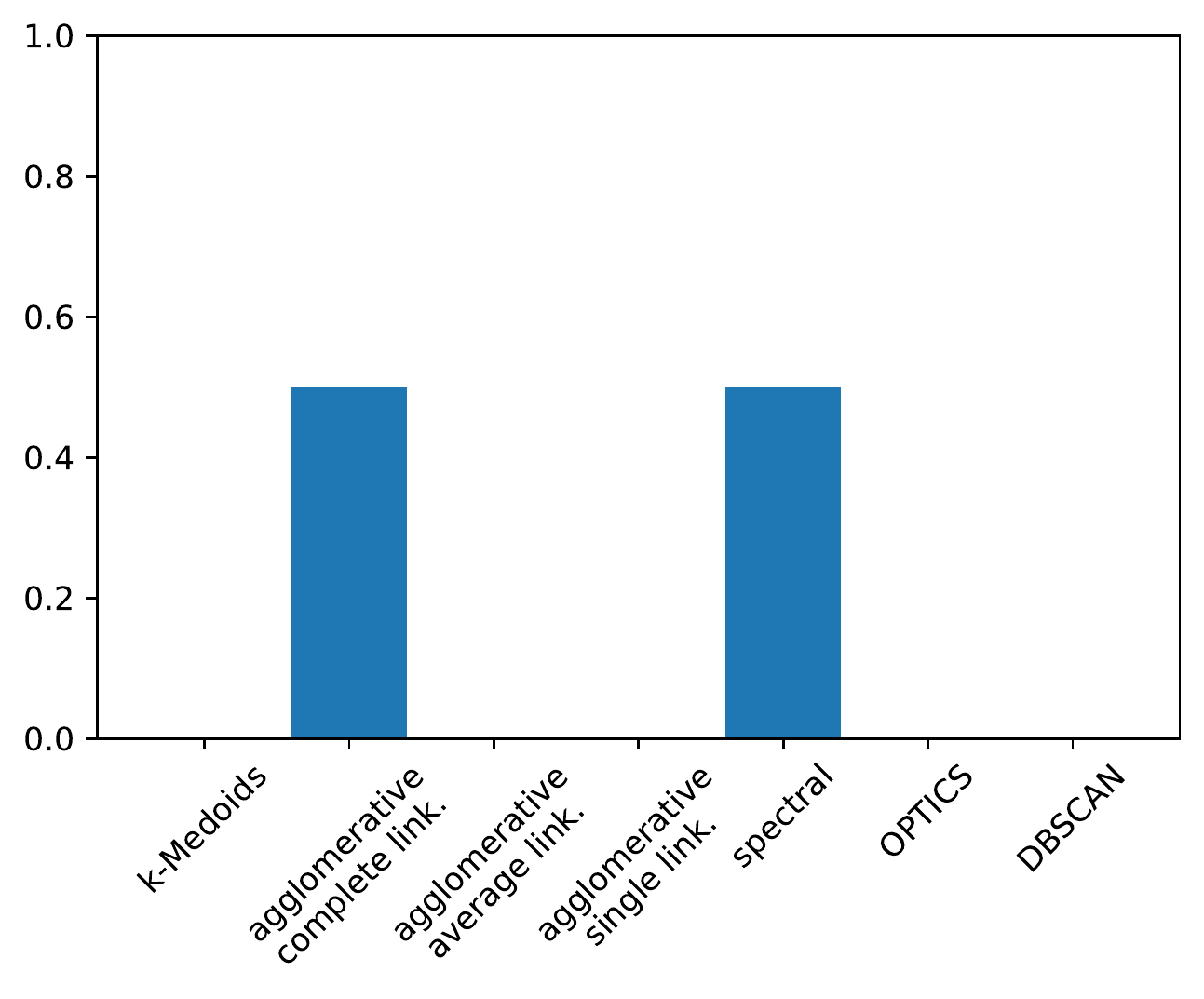}
  \label{fig:inD3_top_freq_algs}
\end{subfigure}
\caption*{inD intersection 2}

\begin{subfigure}{0.4\linewidth}
  \centering
  \includegraphics[width=.9\linewidth]{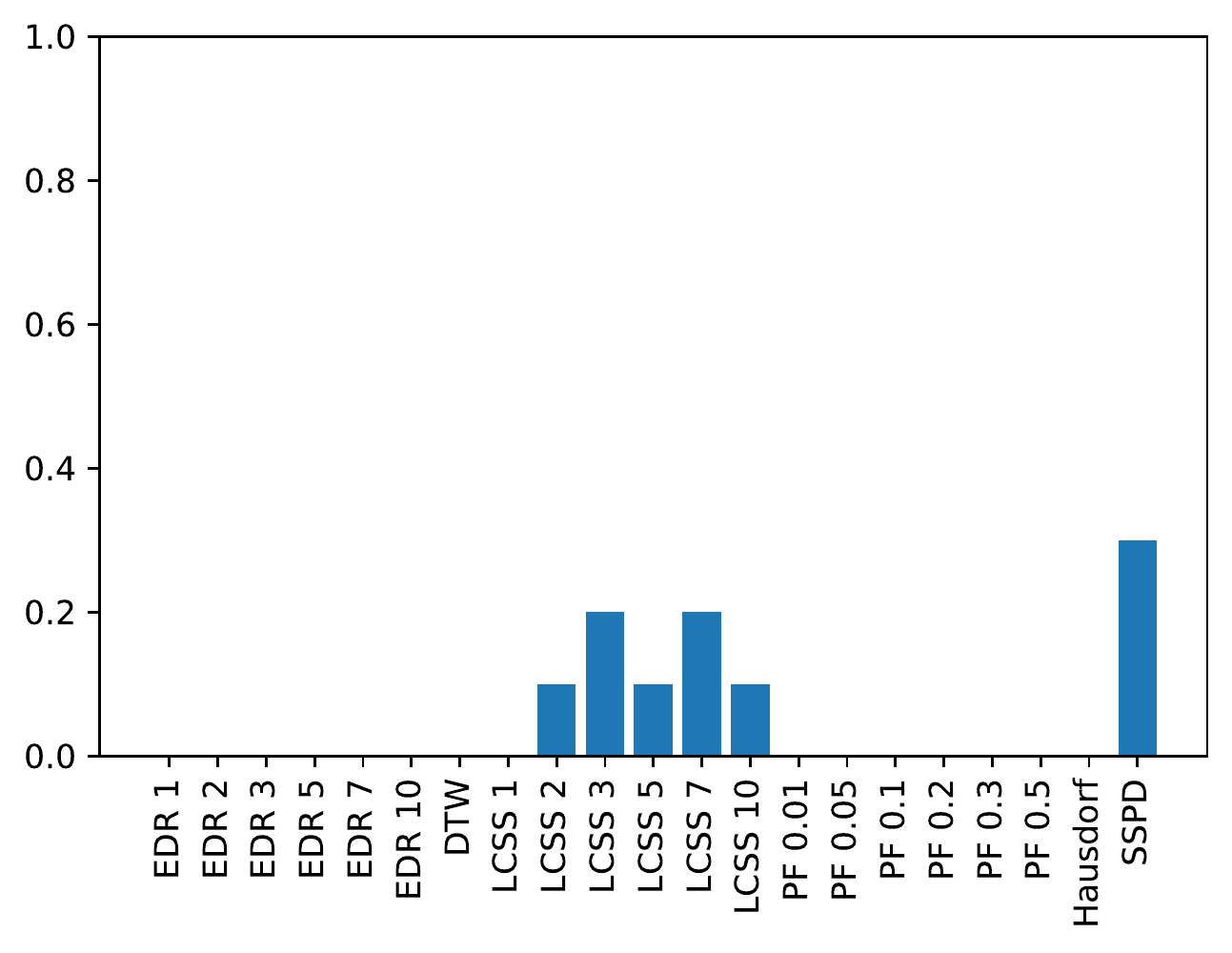}
  \label{fig:inD4_top_freq_simMeasures}
\end{subfigure}
\begin{subfigure}{0.4\linewidth}
  \centering
  \includegraphics[width=.9\linewidth]{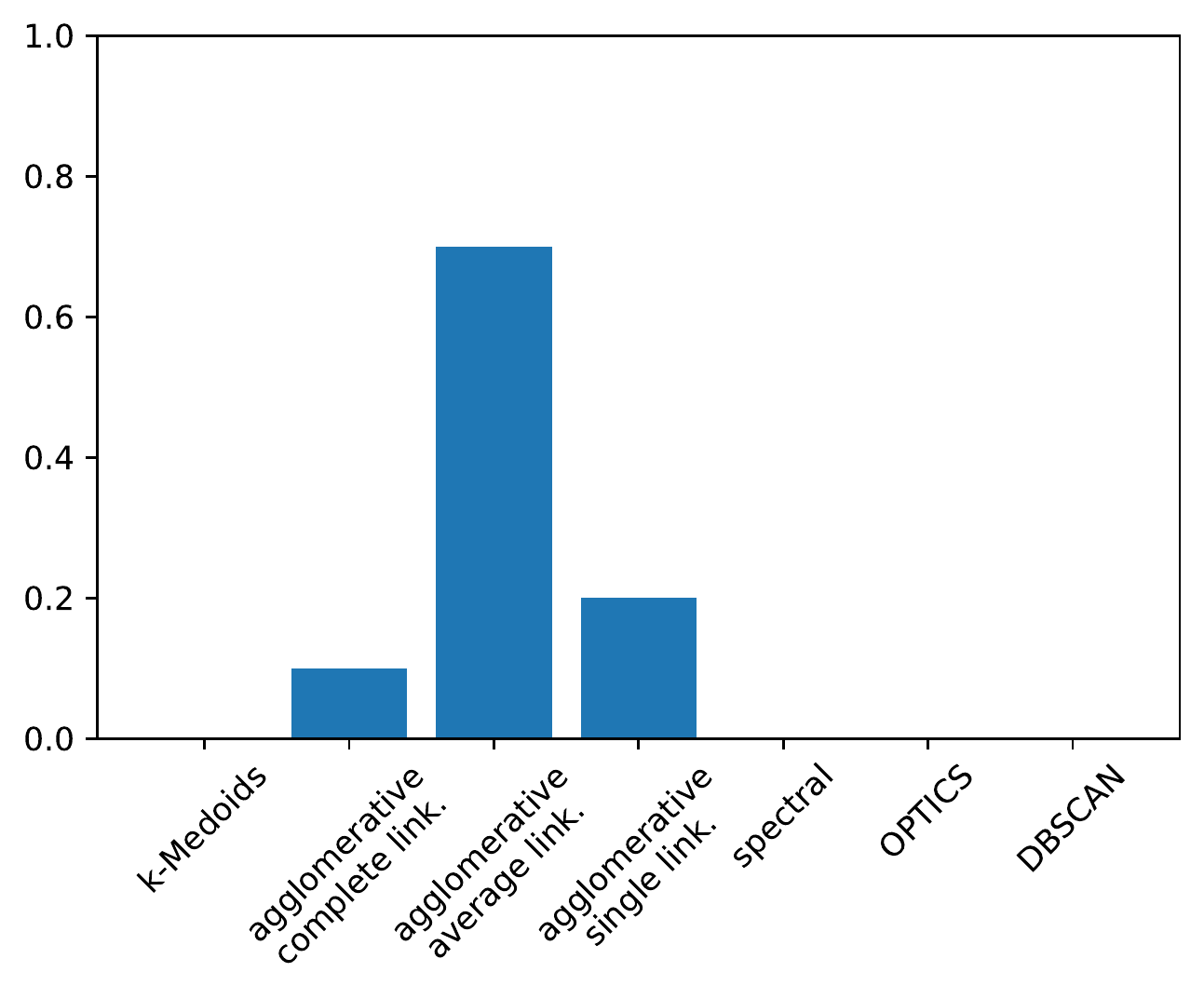}
  \label{fig:inD4_top_freq_algs}
\end{subfigure}
\caption*{inD intersection 3}

\caption{Top performing similarity measures (left) and trajectory clustering algorithms (right) on different \textit{inD} sites}
\label{fig:top_simMeasures_and_algs}
\end{center}
\end{figure}
\begin{figure}[htbp]
\begin{center}

\begin{subfigure}{0.39\linewidth}
  \centering
  \includegraphics[width=.9\linewidth]{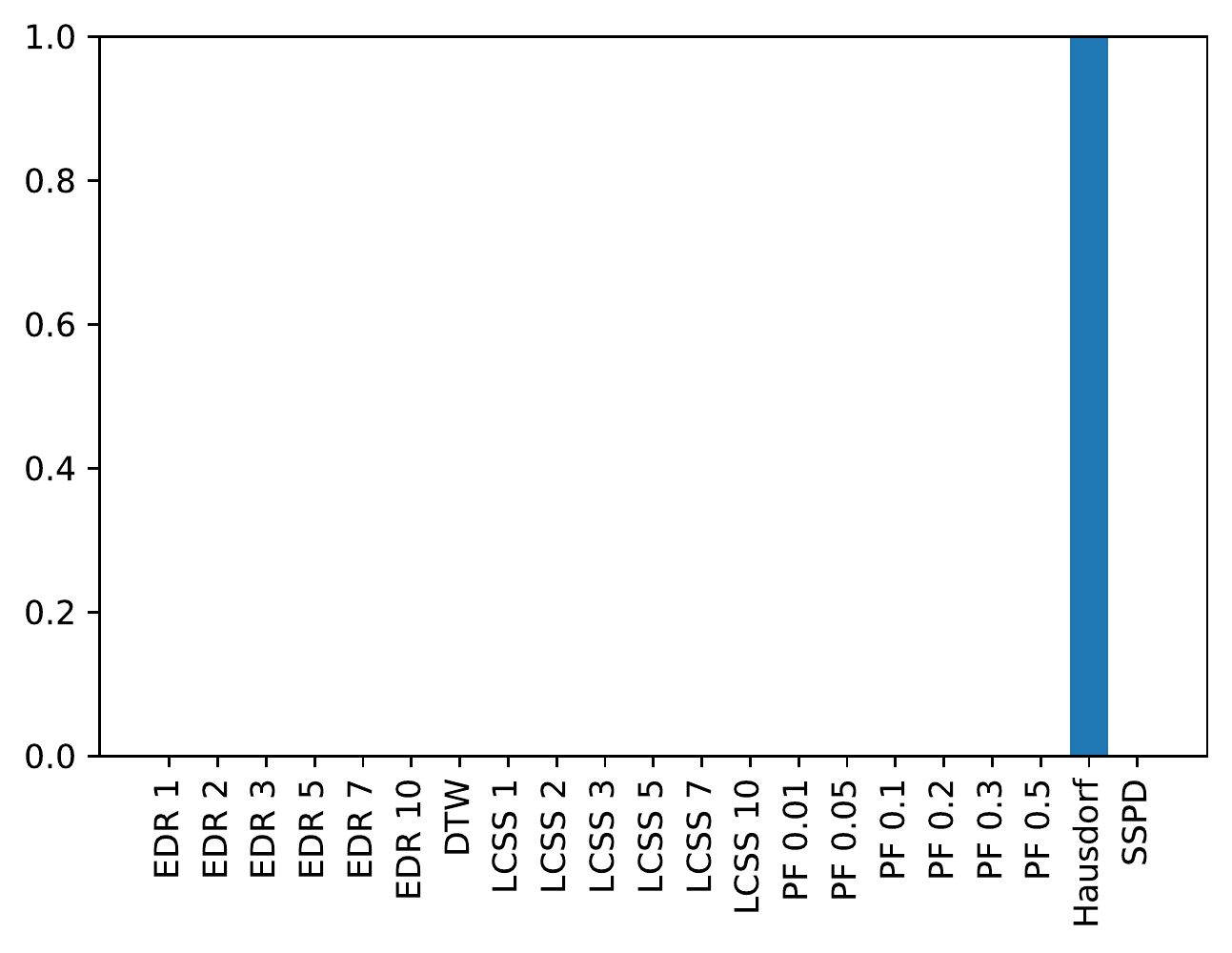}
  \label{fig:NGSIM1_top_freq_simMeasures}
\end{subfigure}
\begin{subfigure}{0.39\linewidth}
  \centering
  \includegraphics[width=.9\linewidth]{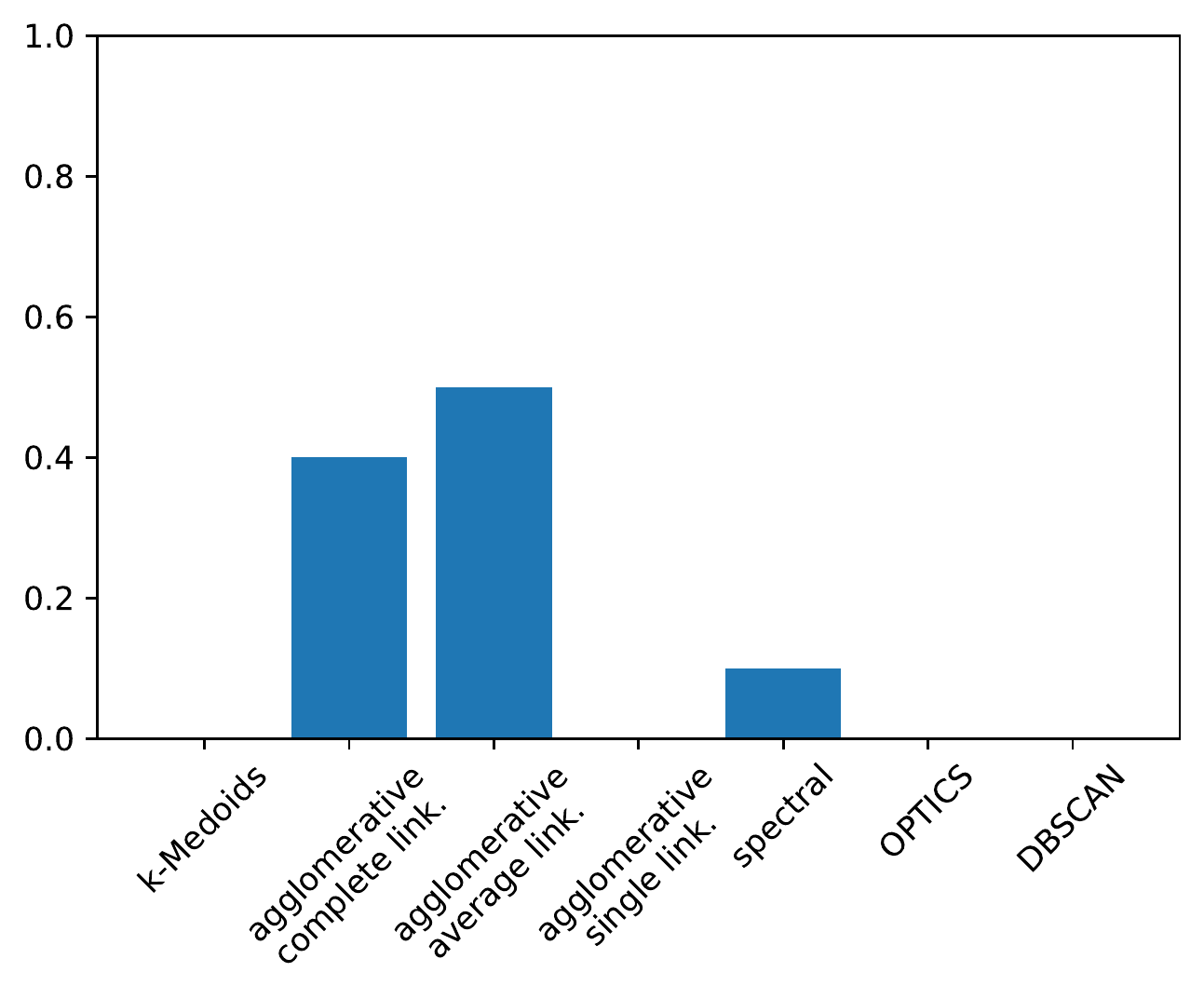}
  \label{fig:NGSIM1_top_freq_algs}
\end{subfigure}
\caption*{NGSIM intersection 1}

\begin{subfigure}{0.39\linewidth}
  \centering
  \includegraphics[width=.9\linewidth]{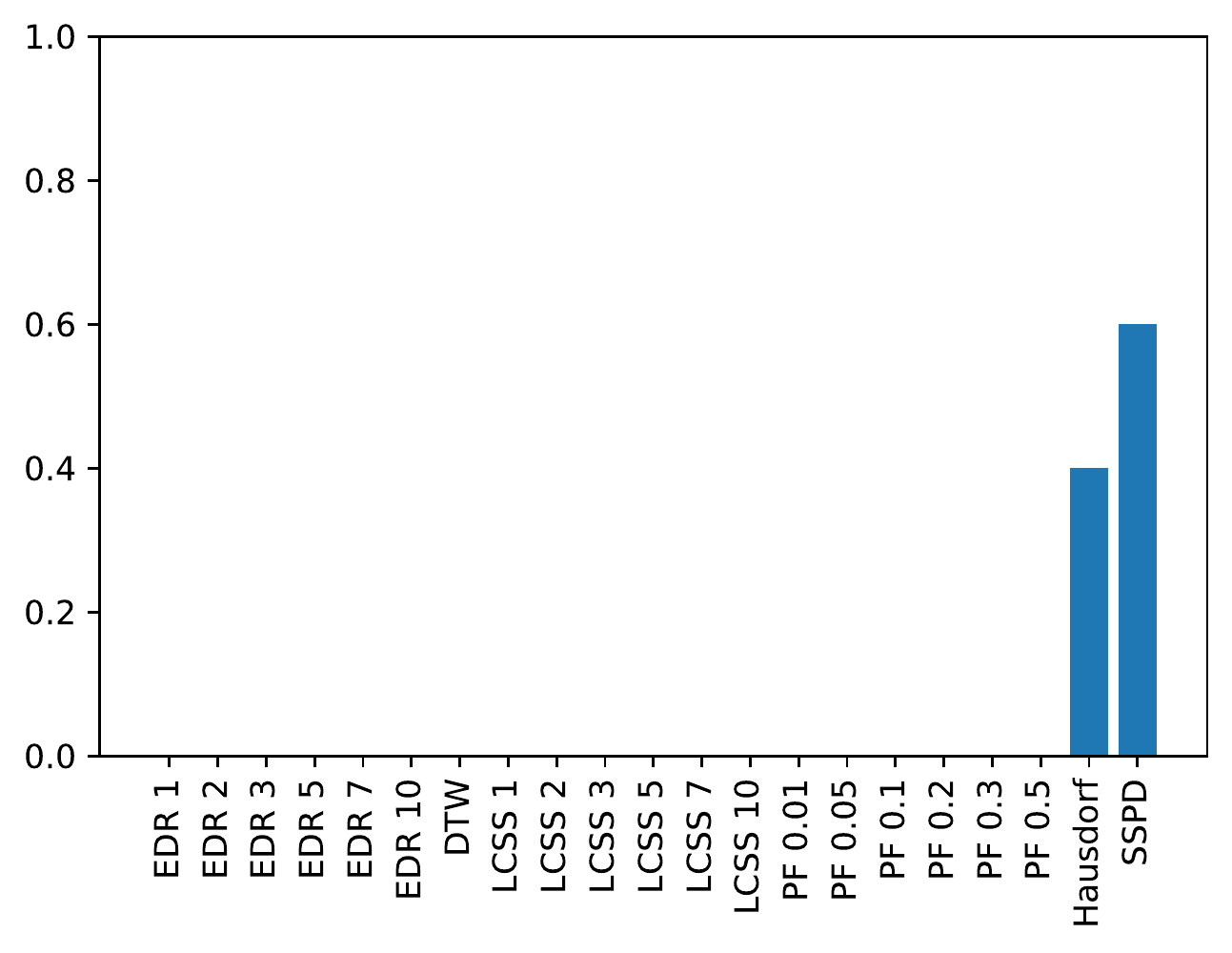}
  \label{fig:NGSIM2_top_freq_simMeasures}
\end{subfigure}
\begin{subfigure}{0.39\linewidth}
  \centering
  \includegraphics[width=.9\linewidth]{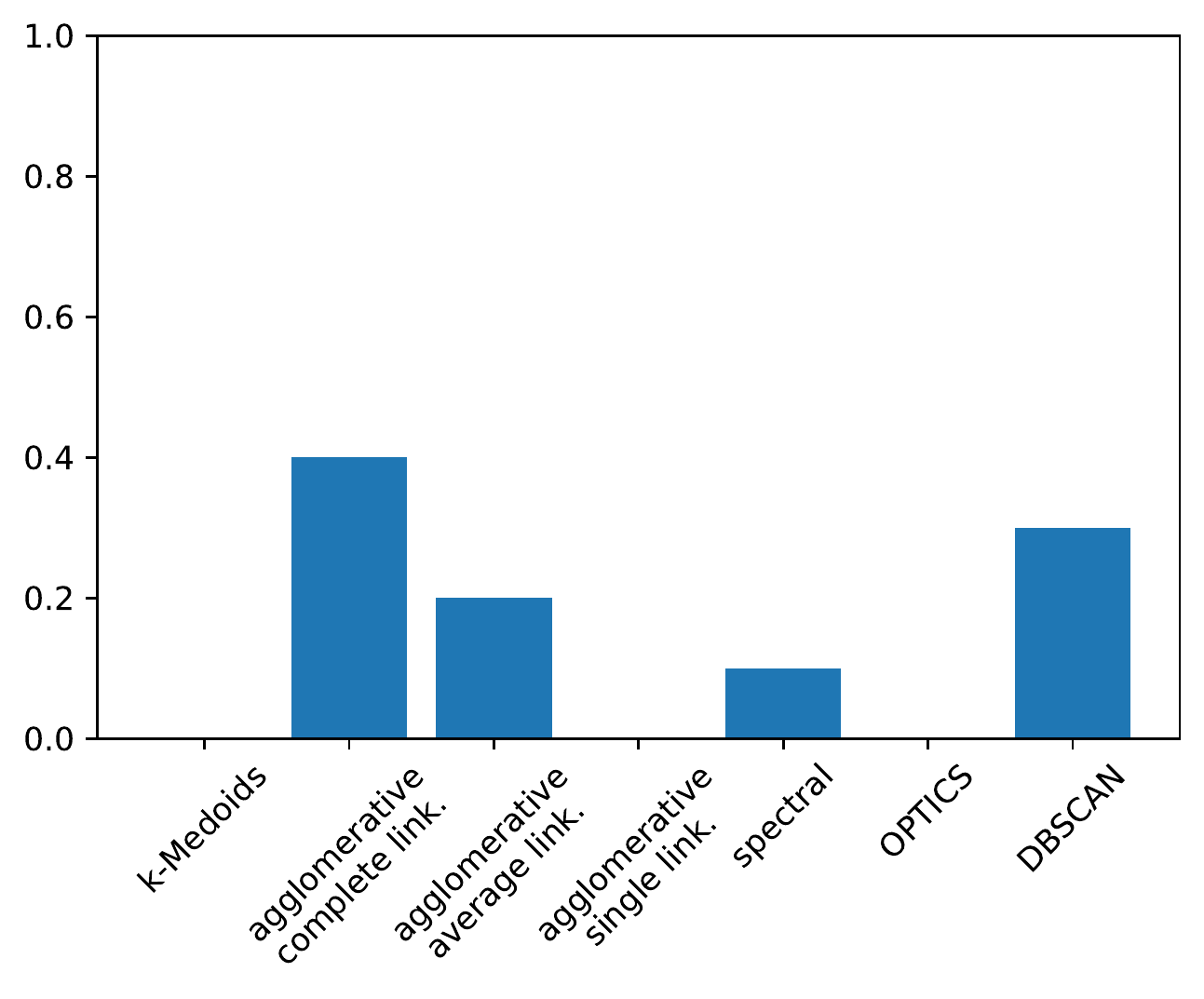}
  \label{fig:NGSIM2_top_freq_algs}
\end{subfigure}
\caption*{NGSIM intersection 2}

\begin{subfigure}{0.39\linewidth}
  \centering
  \includegraphics[width=.9\linewidth]{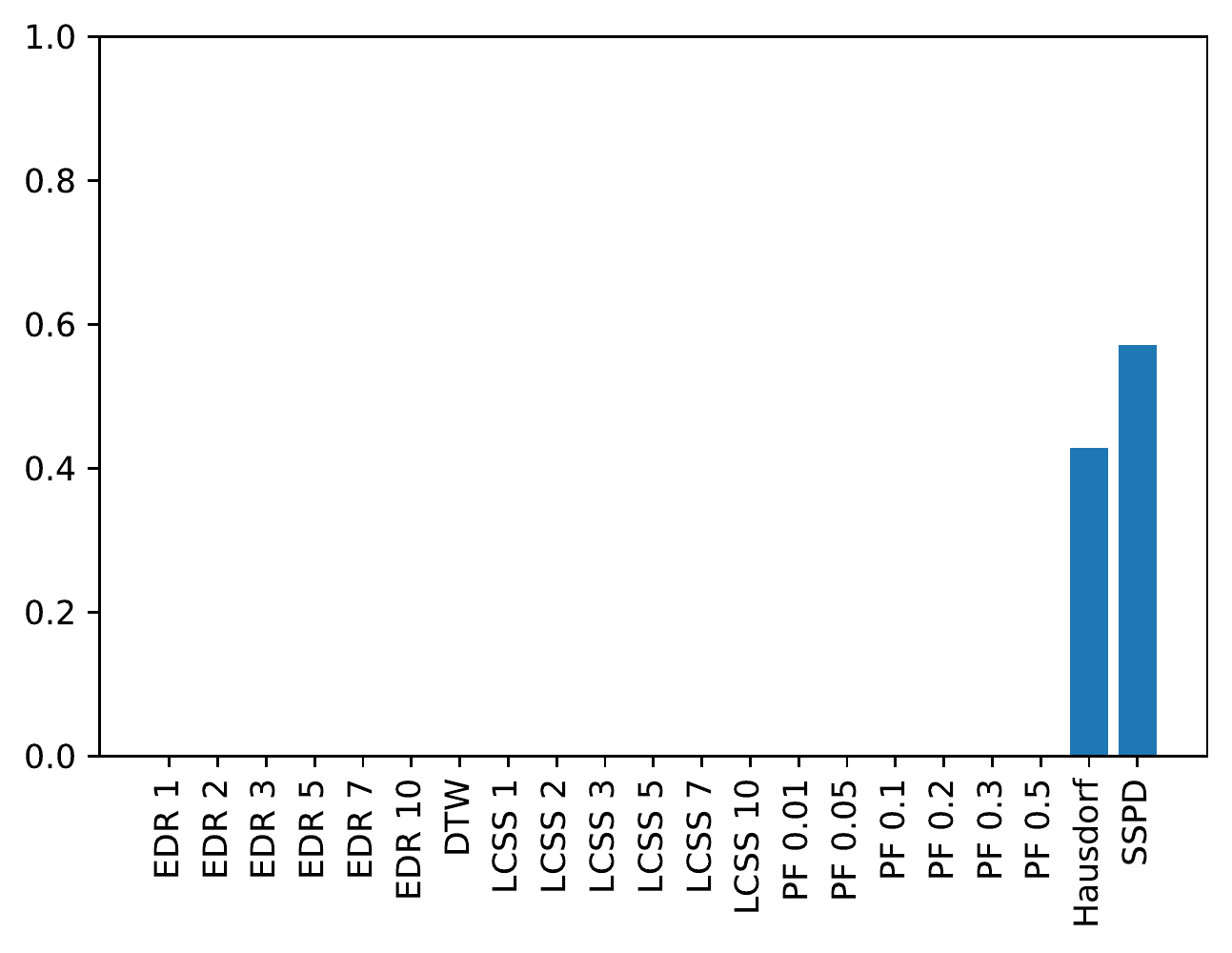}
  \label{fig:NGSIM3_top_freq_simMeasures}
\end{subfigure}
\begin{subfigure}{0.39\linewidth}
  \centering
  \includegraphics[width=.9\linewidth]{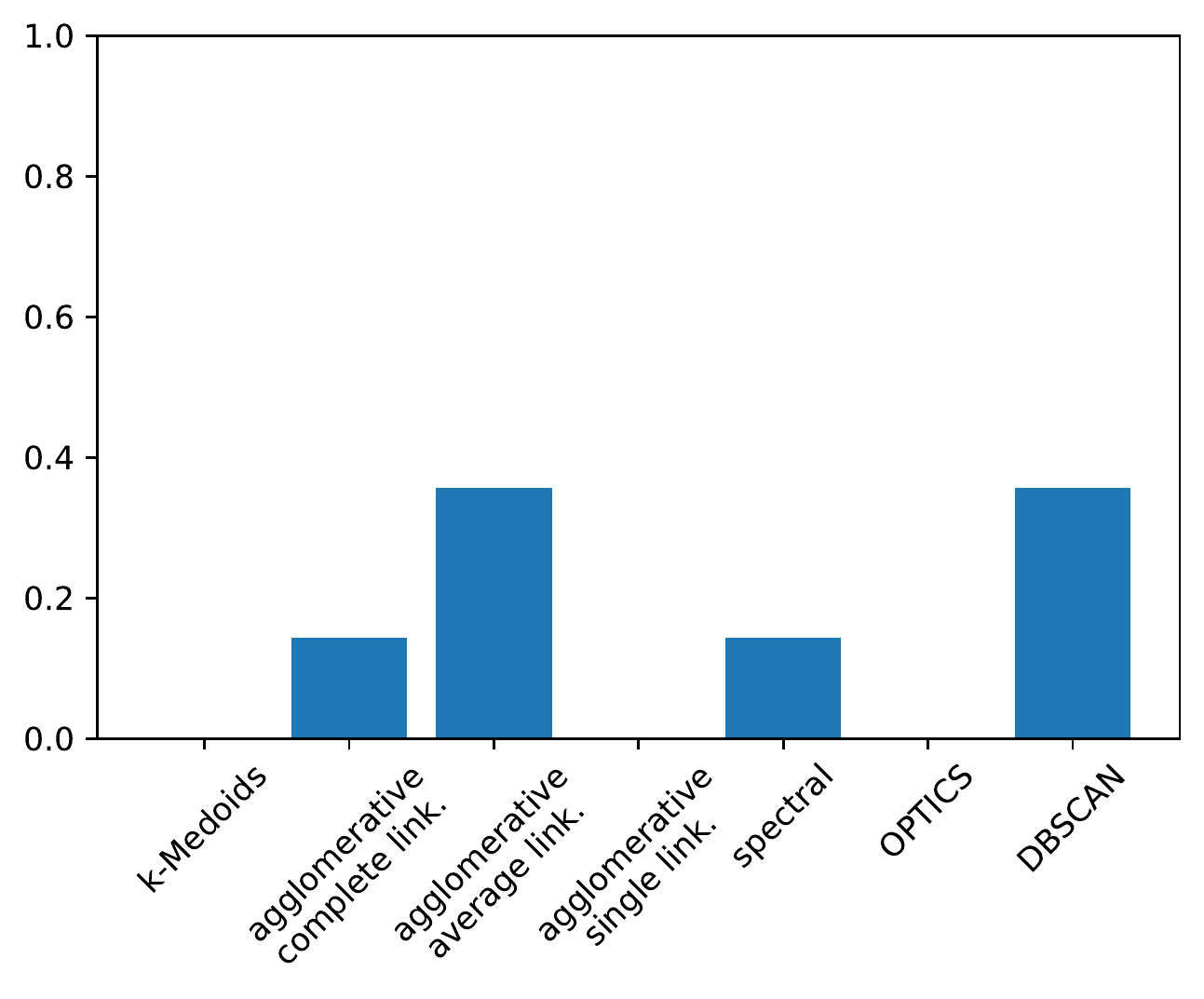}
  \label{fig:NGSIM3_top_freq_algs}
\end{subfigure}
\caption*{NGSIM intersection 3}

\begin{subfigure}{0.39\linewidth}
  \centering
  \includegraphics[width=.9\linewidth]{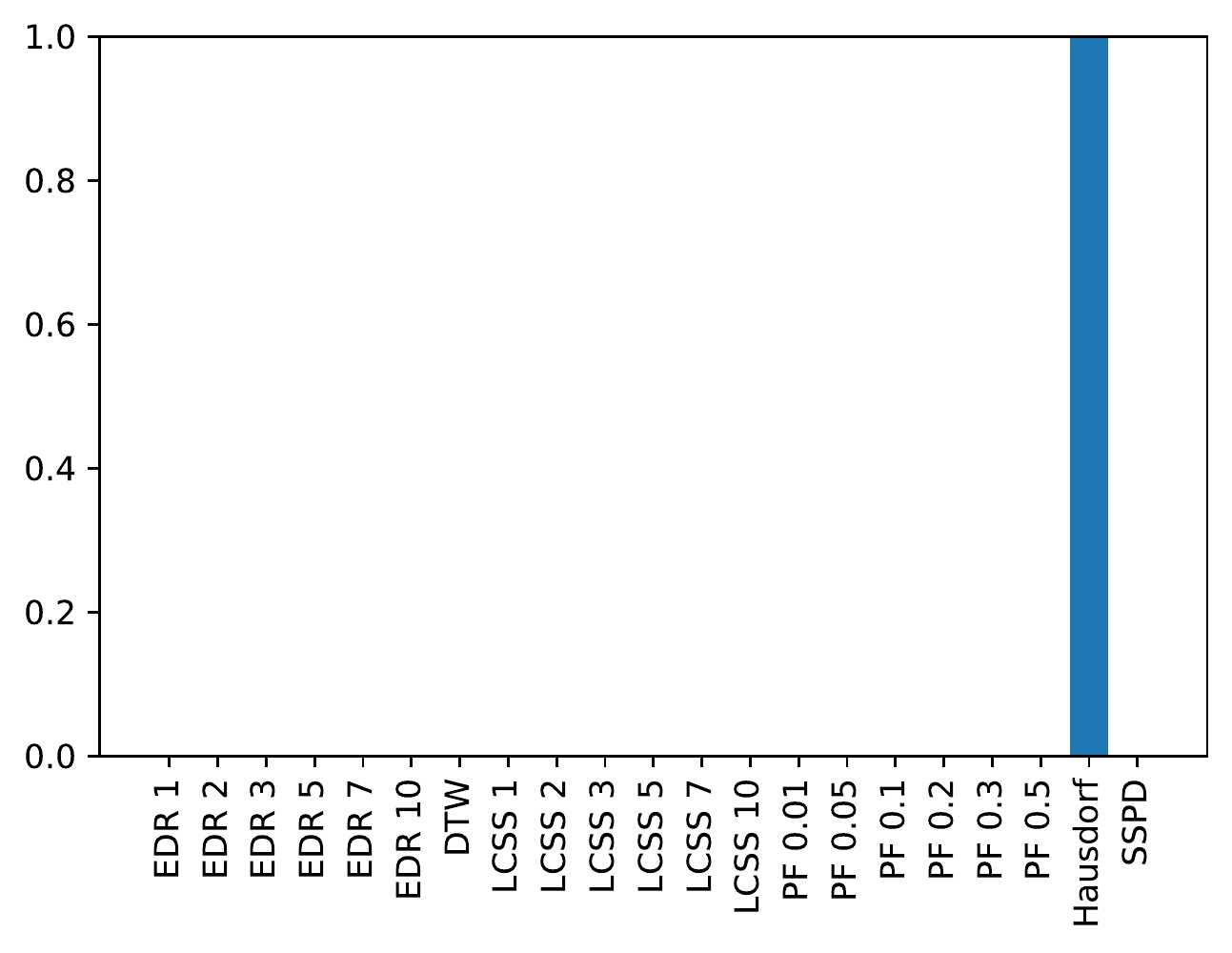}
  \label{fig:NGSIM5_top_freq_simMeasures}
\end{subfigure}
\begin{subfigure}{0.39\linewidth}
  \centering
  \includegraphics[width=.9\linewidth]{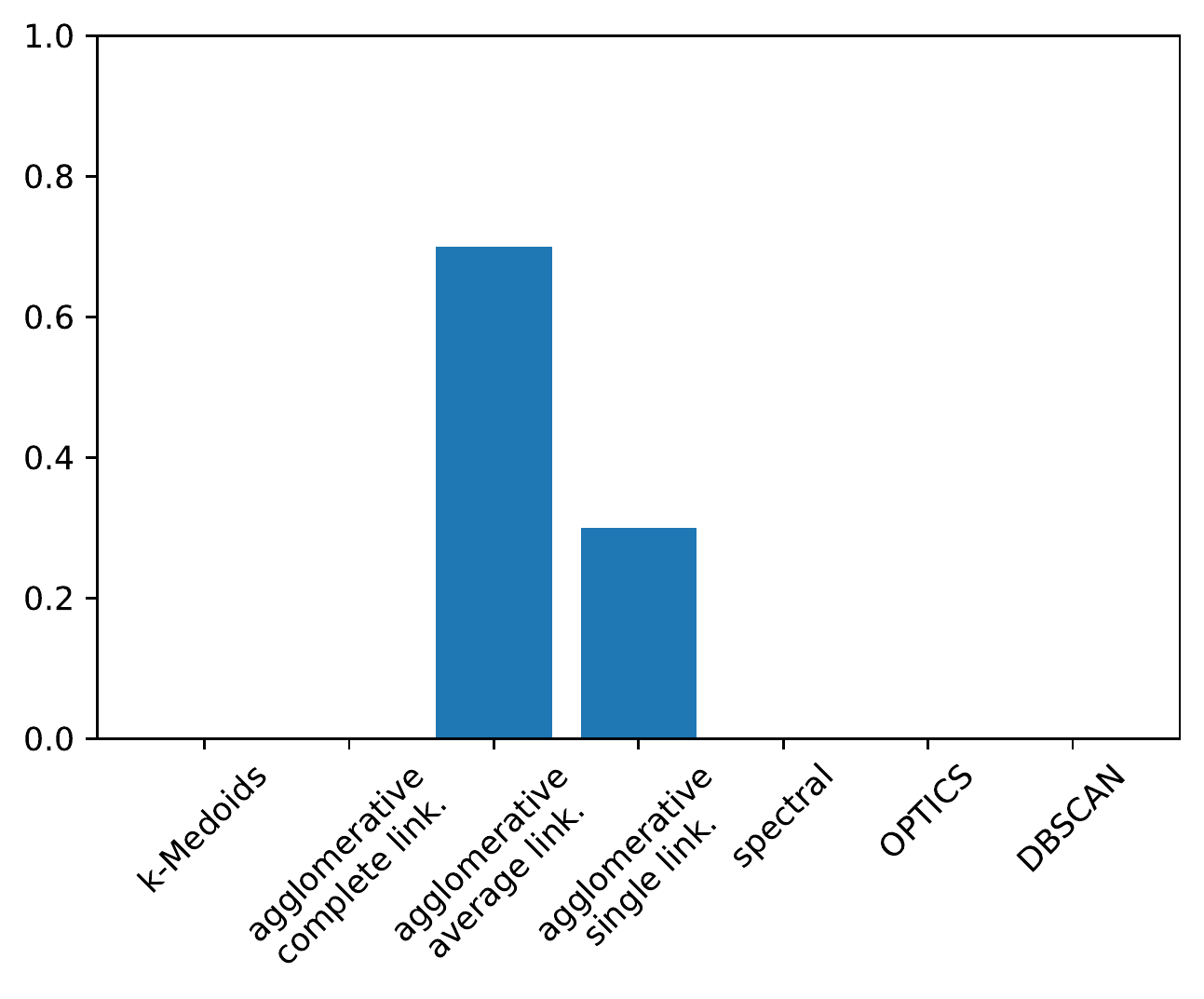}
  \label{fig:NGSIM5_top_freq_algs}
\end{subfigure}
\caption*{NGSIM intersection 4}

\caption{Top performing similarity measures (left) and trajectory clustering algorithms (right) on different 
\textit{NGSIM} 
sites}
\label{fig:top_simMeasures_and_algs_continued}
\end{center}
\end{figure}

The ten best setups for each site are picked based on the average rank $\bar{r}_\omega$. The Figures~\ref{fig:top_simMeasures_and_algs} and~\ref{fig:top_simMeasures_and_algs_continued} shows the top performing distances and clustering algorithms for each site, more precisely the proportion of the 10~best setups in which each distance and algorithm appears. The first observation is that there is no one clustering setup and not even a specific distance measure or clustering algorithm that outperforms systematically the others or at least always appears among the top ten best clustering setups for all sites. Hierarchical clustering is the most common clustering algorithms appearing in the top ten for all sites. Though not similarly frequent, the spectral clustering algorithm is also showing some consistency in appearing among the top performers in most, although not all, the sites. For distances, SSPD and Hausdorff appear frequently, especially for the NGSIM datasets, while the LCSS appears in the top ten for three inD datasets. 

The second observation is that there are clear winners for specific sites. Each site has one, up to three, algorithms or distances outperforming the others (in some cases, LCSS or hierarchical clustering appear several times in the top with different parameters). However, there is no clear pattern in these observations across sites, whether some dataset characteristics may explain to some extent the heterogeneity or whether trajectory clustering is intrinsically sensitive to the input data.  

%% file: 9-6-conc.tex
\section{Conclusion}\label{ch:paper_conclusion}

In this paper, we proposed a way to combine unsupervised and supervised clustering performance metrics without any manual annotation of the trajectories. 
Namely we tried six distances, three of them each tried with six parameter values, i.e.\ a total of 21 distances, six clustering algorithms, different numbers of clusters when used as input by some the clustering algorithms ranging from 2 to 30 and used seven performance measures. We did the experiments on seven intersections from two different datasets, collected in different countries, different road types and through different data collection methods.

Even though generally useful, unsupervised performance measures use the same distance $d$ used in the clustering, and their results often align with the clustering results
As complementary performance measures,
we proposed a method to generate reference clusters based on the origins and destinations of the trajectories to be used in supervised performance measures.

Finally, using a metric based on a combination of performance measures, we picked the top performing distances and clustering algorithms. The results show that there is no single combination of distance and clustering algorithm that is always among the top ten clustering setups. There are distances that performed better compared to the others in specific intersections, but none of them are present among the top performers for every intersection. Among the clustering algorithms, agglomerative hierarchical clustering was the only one that was among the top performers in all seven intersections, though not always with the same linkage type. Also the plots do not show a visual dominance of it over other clustering algorithms. 

Based on these findings, we believe that the choice of similarity measure and clustering algorithm greatly depends on the intersection type, environment and road user behaviors. Therefore, we suggest that similar comparison procedures be followed for new sites among the candidate setups. To facilitate this, we have made all the essential codes developed in this work available online under an open source license (\href{https://github.com/Mr28/Poly-MSc-thesis-trajectory-clustering}{GitHub repository}).

A systematic search of the parameters of the clustering algorithms and distances is needed to determine whether some results depend on the choice of parameters. Further work is necessary to better characterize clustering performance, e.g.\ based on domain specific tasks like origin-destination clustering, and apply these to more datasets and better characterize the factors that determine clustering performance. 
